\documentclass[sigconf]{acmart}

\usepackage{booktabs} % For formal tables
\usepackage{graphicx}
\usepackage{algorithm}
\usepackage{algpseudocode}
\usepackage{bm}
\usepackage{balance}
\usepackage{subfigure}

\def\BibTeX{{\rm B\kern-.05em{\sc i\kern-.025em b}\kern-.08emT\kern-.1667em\lower.7ex\hbox{E}\kern-.125emX}}

\newtheorem{assumption}{Assumption}

\newcommand{\indep}{\perp \!\!\! \perp}

\begin{document}

\title{Markdowns in E-Commerce Fresh Retail: A Counterfactual Prediction and Multi-Period Optimization Approach}

\author{Junhao Hua, Ling Yan, Huan Xu, Cheng Yang}
\affiliation{
	\institution{Alibaba Group}
	\city{Hangzhou} 
	\state{Zhejiang} 
	\country{China}
	\postcode{311121}
}
\email{{junhao.hjh, yanling.yl, huan.xu, charis.yangc}@alibaba-inc.com}
\renewcommand{\shortauthors}{Hua et al.}

\begin{abstract}
	While markdowns in retail have been studied for decades in traditional business, nowadays e-commerce fresh retail brings much more challenges. Due to the limited shelf life of perishable products and the limited opportunity of price changes, it is difficult to predict sales of a product at a counterfactual price, and therefore it is hard to determine the optimal discount price to control inventory and to maximize future revenue. Traditional machine learning-based methods  have high predictability but they can not reveal the relationship between sales and price properly. Traditional economic models have high interpretability but their prediction accuracy is low. In this paper, by leveraging abundant observational transaction data,  we propose a novel data-driven and interpretable pricing approach for markdowns, consisting of counterfactual prediction and multi-period price optimization. Firstly, we build a semi-parametric structural model to learn individual price elasticity and predict counterfactual demand. This semi-parametric model takes advantage of both the predictability of nonparametric machine learning model and the interpretability of economic model.	Secondly, we propose a multi-period dynamic pricing algorithm to maximize the overall profit of a perishable product over its finite selling horizon. Different with the traditional approaches that use the deterministic demand, we model the uncertainty of counterfactual demand since it inevitably has randomness in the prediction process. Based on the stochastic model, we derive a sequential pricing strategy by Markov decision process, and design a two-stage algorithm to solve it. The proposed algorithm is very efficient. It reduces the time complexity from exponential to polynomial. Experimental results show the advantages of our pricing algorithm, and the proposed framework has been successfully deployed to the well-known e-commerce fresh retail scenario - Freshippo.
\end{abstract}




% Copyright

\copyrightyear{2021} 
\acmYear{2021} 
\setcopyright{acmcopyright}
\acmConference[KDD '21]{The 27th ACM SIGKDD Conference on Knowledge Discovery and Data Mining}{August 14--18, 2021}{Virtual}
\acmBooktitle{The 27th ACM SIGKDD Conference on Knowledge Discovery and Data Mining (KDD '21), August 14--18, 2021, Virtual}
%\acmPrice{15.00}
%\acmDOI{10.1145/3292500.3330700}
%\acmISBN{xxx-x-xxxx-XXXX-X/xx/xx}
\fancyhead{}

\maketitle

\section{Introduction}
In a fresh retail scenario, the freshness of goods is the main concern of the customers. Many perishable products, such as vegetables, meat, milk, eggs, bread, have a limited shelf life. To provide fresh and high-quality goods, it is very important to control inventory. If goods have not been sold out before their expiry date, the retailer will have a substantial loss. To maximize the total profit, promotional markdown is a common approach for e-commerce fresh retails.   However,  typically, the retailer does not know what is the best discount price and often chooses an empirical discount on goods (such as $30\% $ off, $50\%$ off),  which is usually not the optimal solution.

\begin{figure}[tbp!] 
	\centering   
	\subfigure[Normal channel]
	{
		\label{fig:normal_channel}
		\includegraphics[width=0.45\linewidth]{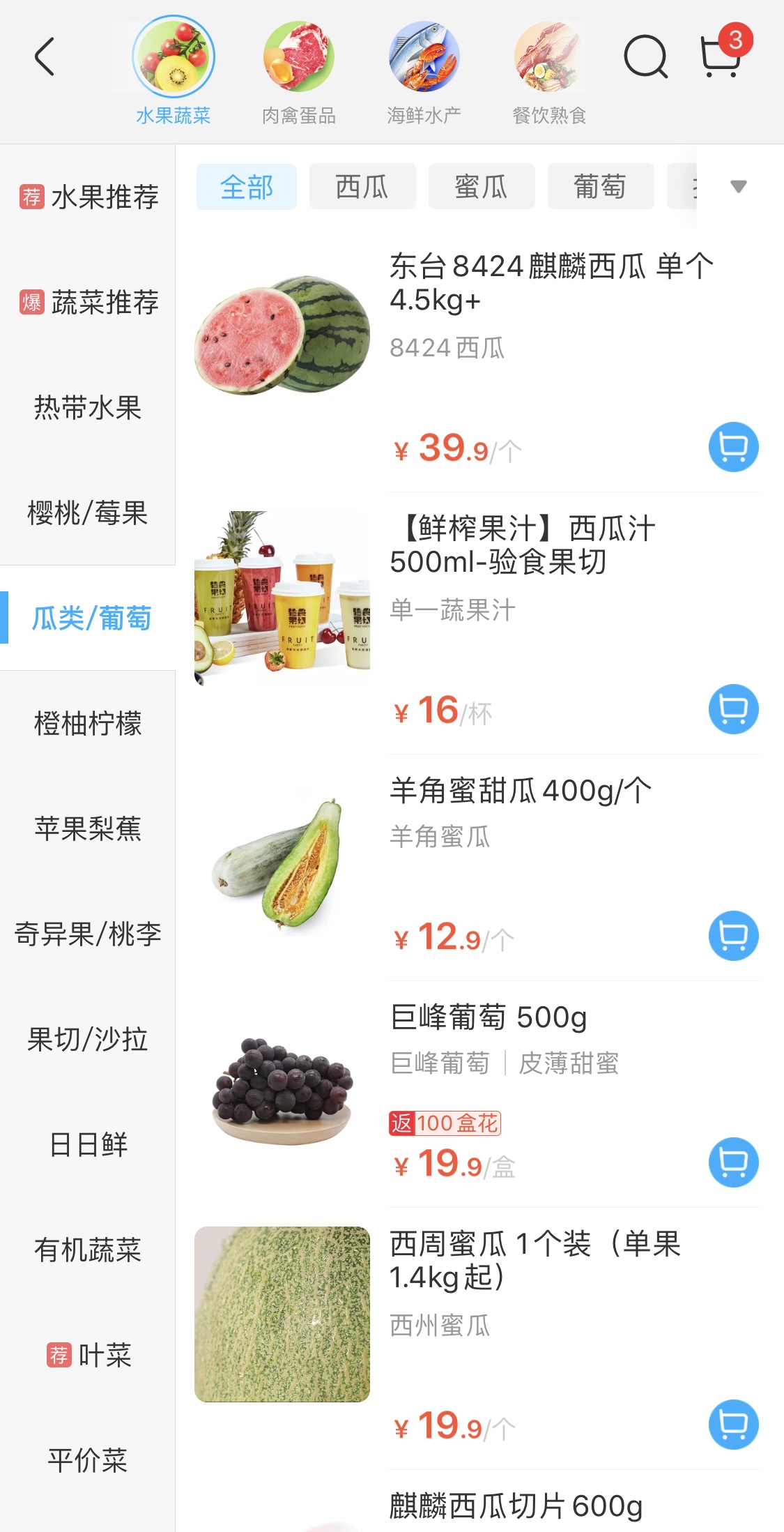}
	}
	\subfigure[Markdown channel]
	{
		\label{fig:markdown_channel}
		\includegraphics[width=0.45\linewidth]{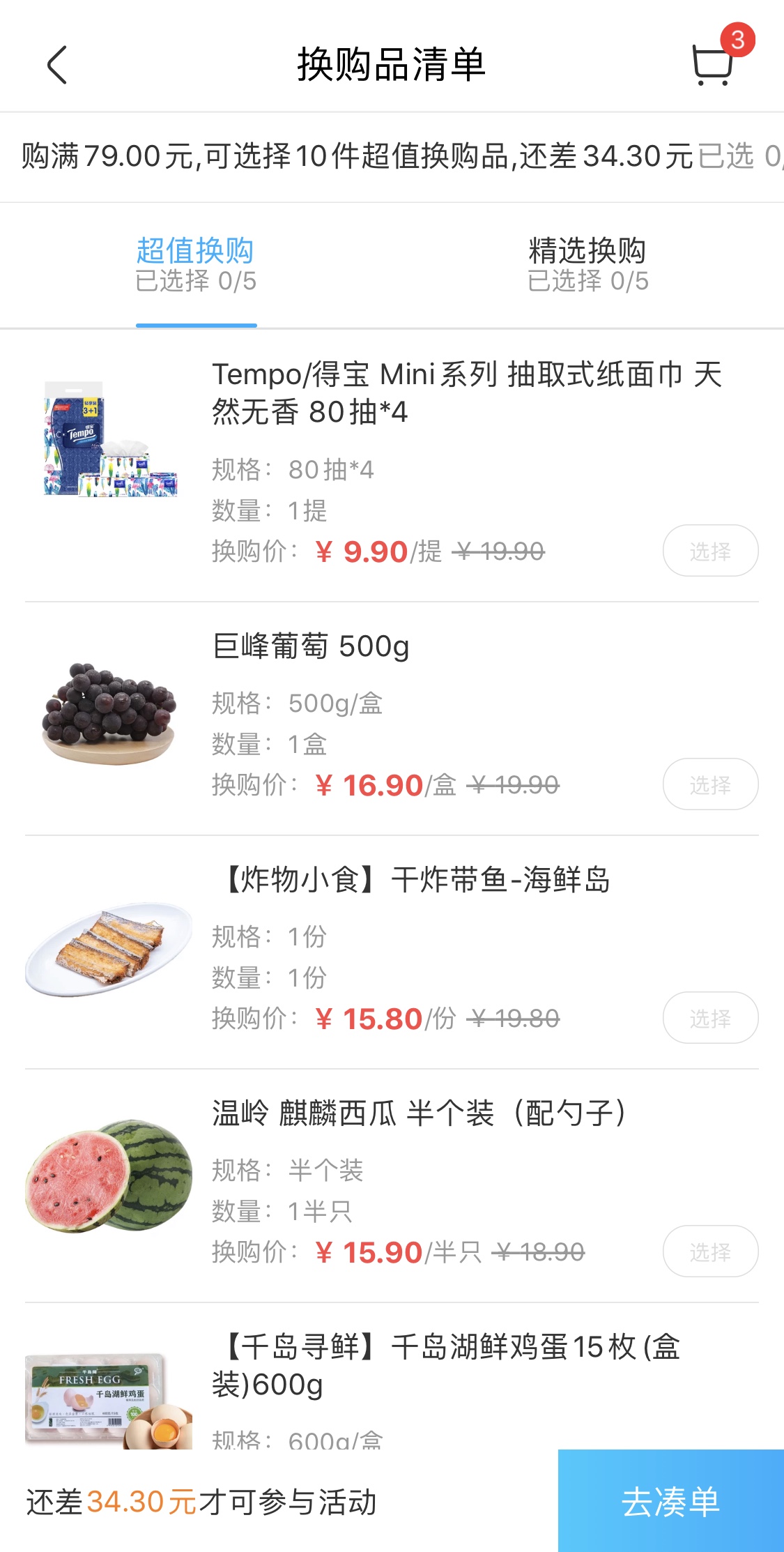}
	}
	\caption{An example of normal channel and markdown channel on Freshippo.} 
	\label{fig:hema}
\end{figure}

In this paper, we consider the e-commerce retailer that its fresh store has two channels for selling goods: the \emph{normal channel}, where goods are sold by no-discount retail price,  and the \emph{markdown channel}, where customers can buy goods by discount under the condition that their total purchase has reached a certain amount.  In particular, we consider the well-known e-commerce fresh retail - Freshippo\footnote{Freshippo (https://www.freshhema.com/) is an online-to-offline (O2O) service platform provided by Alibaba group. It takes the online shopping experience to bricks-and-mortar stores.  It integrates the online and offline stores, uses artificial intelligence technology and operations research methods to provide customers best shopping experience and build ecological industrial chain. }. As shown in \figurename{ \ref{fig:hema}}, customers can either buy goods in normal channel with the retail price or buy them in markdown channel with the discount price. In order to maximize the profit, the retailer needs to answer two questions. First, can goods be sold out with the retail price before its expiry date? Second, if not, what is the optimal discount price for promotional markdown to ensure the goods being sold out while maximizing the profit? The first problem is about sales forecasting \cite{fan2017product,ma2016demand}, and the second problem is about price-demand curve fitting and price optimization \cite{ferreira2016analytics, fisher2018competition, elmachtoub2017smart, perakis2006competitive}. This paper focuses on  the second problem. Specifically, we consider the multi-period dynamic pricing problem of a perishable product  sold in markdown channel over its shelf life.  Our aim is to learn price-demand model rightly and optimize price over a multi-period time horizon dynamically.

The main challenge of demand learning is that the price of most of goods are changed infrequently for fairness, and some even never be changed.  It is hard to predict the sales of a product at the discount price which is not observed before. In the literature of causality, this problem is called counterfactual inference \cite{pearl2009causal,pearl2009causality,morgan2015counterfactuals}, where the price is the treatment/intervention 
% \footnote{We use the word ``treatment'' and ``intervention'' interchange.} 
of the markdown and the sales is the outcome of the markdown. We are interested in studying how the outcome changes with different values of the intervention. For example, suppose we observe the sales of a product with price A and B, we aim to predict the sales of a product with price C, which is counterfactual. The seller can use the price experimentation for demand learning, however the randomized control trail is too costly and has a risk of price discrimination. Alternatively, we focus on the observational study for learning price-demand curve. In other words, we attempt to learn price-demand function from the observational data. It is infeasible to learn individual demand function of a product using only its own data due to the limited price change. Nevertheless, it is possible to jointly learn the price-elasticity of plenty of products with shared parameters. Therefore, we collect abundant feature and historical transaction data of all kinds of products, possibly in different stores. To learn the demand function, a naive approach is using machine learning-based predictive model. It treats price as one of input features and treats  sales as label, and fits the model by minimizing factual error.  However, the small factual error does not mean a small counterfactual error since the  counterfactual distribution may be very different with factual distribution \cite{shalit2017estimating, louizos2017causal}. Besides, when features are correlated with the price (such as historical sales),  the importance of price will be largely dominated. Moreover, since most of the machine learning models are complex or even black-box, it is difficult to reveal the true relationship between price and demand. Therefore, most of ML models have weak interpretability. 

To tackle with this problem meanwhile making use of rich features, we propose a novel \emph{data-driven semi-parametric  structural model} to capture the price-demand curves. This semi-parametric model establishes a connection between black-box machine learning model and economic model, where the role of ML model is to predict the baseline demand, and the role of economic model is to build a parametric and explainable price-demand relationship,  namely price elasticity. To learn price elasticity of individual product,  we make use of the shared information among products and propose a multi-task learning algorithm. Based on this framework, the learned elasticity is stable and the counterfactual prediction is reliable.

Using the counterfactual prediction results, the aim of price optimization is to choose the best discount under some business constraints to maximize the overall profit. Note that a fresh retail usually has many stores in a region. For example, Freshippo has more than 50 stores in Shanghai. To avoid price discrimination, the discounts of the same product in different stores within the same region should be all equal. Therefore, to optimize the discount price, we need to take all stores in a region into consideration. Besides, since both demand and inventory of a product would be changed with different stages of product life time horizon, the retailer would like to take \emph{dynamic pricing} strategy for markdowns.  Moreover, note that the forecasting results obtained by counterfactual prediction will inevitably have randomness as the variance and bias occurred in the learning procedure.
To improve the robustness of the algorithm, it is better to model the demand uncertainty.
For these reasons, we split the life cycle of a product into multiple periods and optimize the discount price of each period to maximize the overall profit. We present a \emph{multi-period joint price optimization} formulation using Markov decision process. Since the number of the state grows exponentially with the number of stores, traditional dynamic programming method suffers from the curse of dimensionality. To solve this problem, we develop an efficient two-stage algorithm. It reduces the time complexity from exponential to polynomial. 
To demonstrate the effectiveness of our approach, we present both offline experiment  and online A/B testing  at Freshippo. It is shown that our approach has remarkable improvements compared to the manual pricing strategy.

\begin{figure}[tbp!] 
	\centering
	\includegraphics[width=0.9\linewidth]{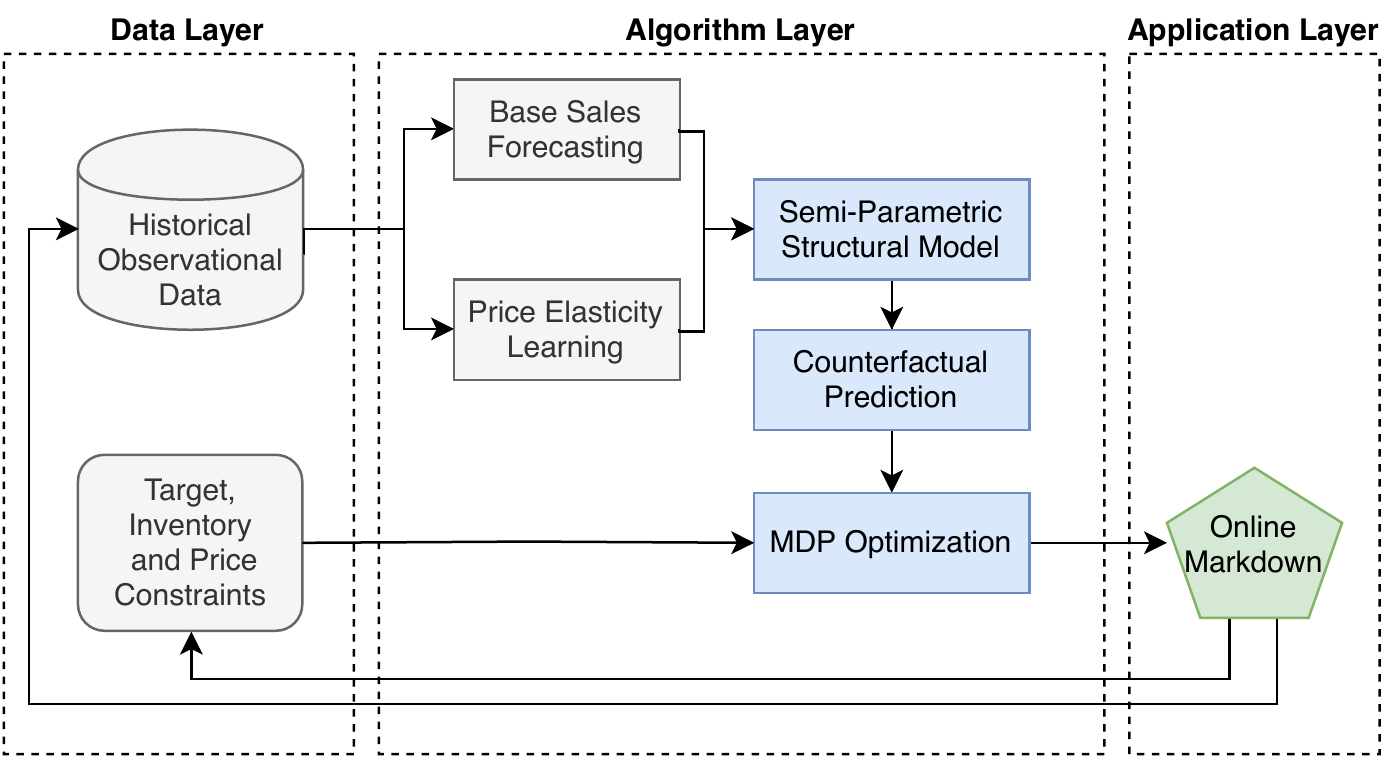} 
	\caption{Our framework for markdowns in fresh retail.} 
	\label{fig:markdown_framework} 
\end{figure}

The main contributions are summarized as follows:
\begin{itemize}
	\item We propose a new learning-and-pricing framework based on counterfactual prediction and multi-period optimization for markdowns of perishable goods, especially for e-commerce fresh retails, as shown in \figurename{ \ref{fig:markdown_framework}}.
	
	\item We propose a \emph{semi-parametric} structural model that taking advantage of both predictability and interpretability for predicting counterfactual demand, and present a \emph{multi-period} dynamic pricing approach which takes the demand uncertainty into consideration, and finally design a very efficient two stage pricing algorithm to solve it.
	
	\item We successfully apply the proposed pricing algorithm into the real-world e-commerce fresh retail. The results show great advantage.
\end{itemize}

\section{Related Work}
This paper addresses the important data science problem, markdown optimization, which focuses on the dynamic pricing problem where the price of a perishable product is adjusted over its finite selling horizon. It is a variant of revenue management (RM), and has been studied in areas of marketing, economics, and operations research \cite{phillips2005pricing}.  
Most of studies address the static price optimization problem using descriptive models
\cite{van2005models,blattberg2010shrinkage,bolton2003empirically}.  We focus on multi-period dynamic pricing optimization with demand uncertainty using the prescriptive model.

In the literature, some researchers have developed pricing decision support tools for retailers. For example, a multiproduct pricing tool is implemented for fast-fashion retailer Zara for its markdown decisions during clearance sales  \cite{caro2012clearance}.  For recommending promotion, the researchers present randomized pricing strategies by incorporating some new features into electronic commerce \cite{wu2014randomized}. In \cite{ferreira2016analytics}, the researchers develop a multiproduct pricing algorithm that incorporates reference price effects for Rue La La to offer extremely limited-time discounts on designer apparel and accessories.

In the context of machine learning, some regression methods have been applied for demand learning. Most of existing works are based on linear regression with different linear demand models for its ease of use and interpretation, such as strict linear, multiplicative, exponential and double-log model \cite{huang2013demand}. Other works consider the market share of products and use choice model to model demands of products, such as multinomial logit model, nested logit model \cite{li2011pricing}. 
Some researchers use semi- and non-parametric models for demand learning.  In \cite{martinez2006evaluating}, the authors use SVM to evaluate retailer's decisions about temporary price discounts. In \cite{hruschka2006relevance}, the authors use multilayer perceptrons to estimate store-specific sales response functions. However, these methods can not work well when the price of a product is seldom changed in the historical data, which is common in real-world scenario.

Some recent approaches use machine learning methods for causal inference, such as trees \cite{athey2016recursive,athey2019generalized} and deep neural networks \cite{shalit2017estimating, louizos2017causal}. But these methods are designed for binary treatment. In \cite{hartford2017deep}, the authors presents a deep model for characterize the relationship between continuous treatment and outcome variables  in the presence of instrument variables (IV). However, the IV itself is hard to be identified, which limits the scope of its application.

In the literature of price optimization, some researchers consider the large scale pricing problem of multiple products and propose algorithms based on network flow \cite{ito2016large},  semi-definite programming relaxation \cite{ito2017optimization} and robust quadratic programming \cite{yabe2017robust}.  In \cite{zhang2006multi}, the authors consider pricing problem in multi-period setting, and use robust optimization to find adaptable pricing policy. However, this model is designed for monopoly and oligopoly and is not suitable for e-commerce fresh retail scenario.

\section{Problem Formulation}
Let us consider an e-commerce retail who wants to sell out $N$ products by markdowns.  
These products are possibly sold in different stores in a region. Our task is to offer the optimal discount price for these products to maximize the overall profit of all stores. 

Instead of using the absolute value of price, we use the relative value, i.e. the percentage of retail price $d \in [0, 1]$, as the decision variable (treatment/intervention)  of markdowns, as the range of the percentage discount is fixed and finite while the price itself is infinite.  
Thus,  we can jointly learn price elasticity of different products who have different price magnitudes.
Let us denote $Y_i^{obs}$ as the average sales of product  $i$ at the discount $d_i$  in markdown channel in the past days.
We collect a set of observable covariate features $\bm{x}_i \in \mathbb{R}^{n}$, including categories, holidays, event information, inherent properties and  historical sales of products and shops, etc.  In particular, we denote  categorical feature as $L_i \in \{0, 1\}^{m}$, and assume there are three level categories.  

The proposed pricing algorithm consists of two steps: counterfactual prediction and price optimization.  In next two sections, we present these two steps respectively.

\begin{figure}[tbp!] 
	\centering
	\includegraphics[width=0.9\linewidth]{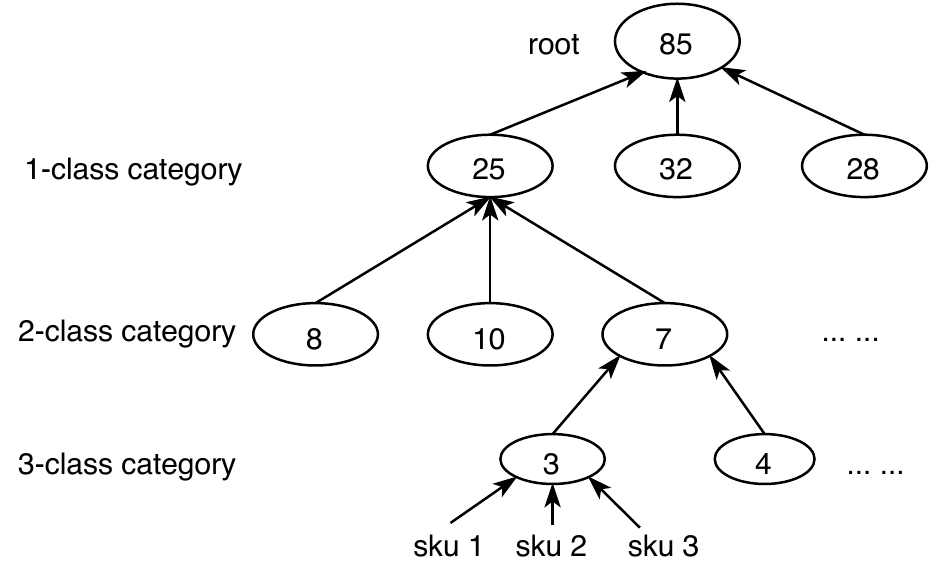} 
	\caption{An example of the data aggregation structure by categories. The number in a circle represents the number of the observed data  aggregated in that category.} 
	\label{fig:category} 
\end{figure}

\section{Counterfactual prediction}\label{sec_demand_learning}
The key of pricing decision making is to accurately predict the demand of products at different discount prices. The difficult is that  these prices may never be observed before.
In other words, they are counterfactual \cite{hofman2017prediction,prosperi2020causal}. The aim of counterfactual prediction is to predict the expectation of demand/sales $\mathbb{E}[Y_i|\text{do}(d_i), \bm{x}_i]$ under the intervention $d_i$ and the condition $X=\bm{x}_i$, where the $\text{do}(\cdot)$ is the do operator \cite{pearl2009causality}.

\subsection{Semi-Parametric Structural Model}

In this subsection, we present a semi-parametric structural model to predict the counterfactual demand. It works under the common simplifying assumption of ``no unmeasured confounding'' \cite{hernan2010causal}. Specifically, we make the following assumption:
\begin{assumption}[unconfoundedness] \label{no_unobserved_confounder}
	The treatment $d_i$ is conditionally independent with potential outcomes $Y(\text{do}(d_i))$ given the observed features $\bm{x}_i$, i.e., $d_i \indep Y_i(\text{do}(d_i)) | \bm{x}_i$.
\end{assumption}
This ensure the identifiability of the casual effect when the complete data on treatment $d_i$,  outcome $Y_i$ and covariates $\bm{x}_i$ are available. However, in the real-world scenarios,
it is almost impossible to collect the complete data for a single product $i$, especially when treatment $d_i$ is continuous. In most cases, the discount on a product only has one or two different values in the historical transaction data. Therefore, it is impossible to fit the price-demand curve and compute individual causal effect using the observational data of a single product. 

To solve this problem, we use the data aggregation technique. We aggregate data of all products by using the \emph{category} information and learn the causal effect of each product \emph{jointly}.
As shown in \figurename{ \ref{fig:category}, a high-level category has many low-level categories, and a lowest-level category has many different products/SKUs (stock keeping unit). Note that a  higher level category has more SKUs, but the difference and variance between SKUs also become larger. 
	
Borrowing the ideas from marginal structural model \cite{robins2000marginal}, we propose a semi-parametric structural model to learn individual casual effect. In detail, we assume the outcome $Y_i$ is structurally determined by $d_i$, $\bm{x}_i$ and $L_i$ as  
\begin{equation}\label{semi_param}
\mathbb{E}[\ln (Y_i/Y_i^{nor}) |d_i, L_i]  = g(d_i; L_i, \bm{\theta}) +  h(d_i^{o}, \bm{x}_i) -  g(d_i^{o}; L_i, \bm{\theta}),
\end{equation}
where $Y_i^{nor}$ is the average sales of product $i$ in normal channel in recent weeks, and $d_i^o$ is the average discount of product $i$  in the markdown channel in recent weeks. We treat $Y_i^{nor}$ as normalized factor, and the quantity $Y_i/Y_i^{nor}$ represents the ratio of the sales with a discount price in markdown channel to the sales with the retail price in normal channel. This quantity is much more useful compared with the absolute value of sales, since sales may have different magnitude with different products.
	
	The function $g(d_i; L_i, \bm{\theta})$ is the parametric price-elasticity model of product $i$ and its parameter $\bm{\theta} \in \mathbb{R}^{m+1}$ represents the price elasticity vector, which is shared by all products. The function $h(d_i^o, \bm{x}_i)$ is nonparametric preditive model whose role is to predict the log ratio of sales  at the base discount $d_i^o$. Note that let $d_i=d_i^o$,  we have
	\begin{equation}
	\mathbb{E}[\ln (Y_i/Y_i^{nor}) |d_i^o] = h(d_i^o, \bm{x}_i).
	\end{equation}
	Model (\ref{semi_param}) is semi-parametric model since it consists of the parametric model $g(\cdot)$ and the non-parametric model $h(\cdot)$. It provides both high interpretability through parametric econometric model and high predictability through nonparametric machine learning model. We respectively present these two models in the following.
	
	\subsection{Base Sales Forecasting} \label{sales_forecast}
	The fundamental building block $h(d_i^o, \bm{x}_i)$ aims to forecast the sales at base discount $d_i^o$ using the observational data $\{ \bm{x}_{i,t}, Y_{i,t}^{obs}\}$, where $t$ represents the index of data points. A naive approach may treat the discount as one of input variables, and train a predictive model using observational data, then predict the counterfactual outcome by varying the discount, i.e., $E[Y_i|\text{do}(d_i)] = h(d_i, \bm{x}_i)$.  However, this method does not work.
	Because transactional features, such as historical sales, are affected by the historical discount. 	
	If we do an intervention on the discount $d_i$, i.e. varying the discount, we also need to change the relevant historical features $\bm{x}_i(d_i)$ which is impossible. To resolve this problem, we dissociate the treatment from the sales-related features, and divide model into two parts
	as presented in model (\ref{semi_param}). We emphasize that the sales forecasting module only predicts the sales at an average of historical discounts $d_i^o$, i.e., 
	\begin{equation} \label{non_param_predictive}
	\begin{split}
	& h(\cdot) \gets \{\bm{x}_{i,t}, Y_{i, t}^{obs}/Y_{i,t}^{nor}\}_{t=1,\dots, T, i=1,\dots, N},   \\
	& \ln (Y_i^{o}/Y_{i, T+1}^{nor})  = h(d_i^o, \bm{x}_{i, T+1}), i=1,\dots, N.
	\end{split}
	\end{equation}
	We call $(d_i^o, Y_i^o)$ as base discount and base sales pair of product $i$. 
	Many fundamental non-parametric predictive model can be applied to learn $h(\cdot)$, such as deep neural networks \cite{lecun2015deep} and boosting tree model \cite{chen2016xgboost}.

\subsection{Price Elasticity Model}
In the second part,  we aim to learn the average price sensitivity of customers to products.
We propose a double-log structural nested mean model, 
\begin{equation}\label{price_elasticity}
g(d_i; L_i, \bm{\theta})  = \mathbb{E}[\ln (Y_i/Y_i^{nor})] =( \theta_1  + \bm{\theta}_2^T L_i ) \ln d_i + c,
\end{equation} 
where $\bm{\theta}_2 \in \mathbb{R}^m$, $\bm{\theta} = [\theta_1, \bm{\theta}_2^T]^T$, $c$ is an intercept parameter and $L_i$ is a compound of three one-hot variables, namely
\begin{equation}
L_i = [\underbrace{0,\dots,1,0}_{category 1},\underbrace{0,1,\dots, 0}_{category 2},\underbrace{0,\dots, 1, 0}_{category 3}]^T.
\end{equation}
Since the coefficient of treatment $d_i$  is modified by $L_i$, we call $L_i$ as effect modifiers.
By exponential transformation, this model can also be written as a constant price elasticity model, i.e.,
\begin{equation}
Y_i =   Y_i^{nor}  e^c d_i^{\theta_1 + \bm{\theta}_2^T L_i }, \forall i,
\end{equation}
where  $\theta_1 + \bm{\theta}_2^T L_i$ is price elasticity of demand, which is structured by three different level categories. In other words, the price elasticity of each individual SKU is a summation of a common coefficient $\theta_1$ and the elasticity of each category. 
As we have stated, the aggregated data within high level category has large sample size
but also has large variation, while the aggregated data within low level category has small variation but  also  has small sample size. Instead of learning elasticity within a single category, this model simultaneously learns price elasticity of all category-levels which is more flexible. 
	
To estimate the price elasticity of demand, we can use the mean squared error criterion with entire samples. However, in the real world e-commerce retail scenario, the data is generated by streaming. A better alternative is to update the parameters in an online fashion. Therefore, based on recursive least squares,  we update the parameters by minimizing the following objective function: 
	\begin{equation} \label{rls_objective}
	\min_{\bm{\theta}, c} \sum_{i=1}^{N} \sum_{j=1}^{t} \tau^{t-j} || \ln \frac{Y_{i,j}}{Y_{i, j}^{nor}} -   \bm{\theta}^T \hat{L}_i \ln d_{i, j} -c ||_2^2 + \lambda ||\bm{\theta}||_2^2,
	\end{equation}
	where $\hat{L}_i= [1, L_i^T]^T$ is the augmented variable and $\lambda>0$ is a pre-defined regularization coefficient, $0<\tau \leq 1$ is a forgetting factor  which gives exponentially less weight to older samples.
	The optimal solution of (\ref{rls_objective}) can be represented by two sufficient statistics, which are then can be easily updated in an online fashion. 
	
\subsection{Counterfactual Demand Prediction}
Suppose we have obtained the observational data from time $1$ to $t-1$, our aim become to predict the counterfactual demand of product $i$ of time $t$ with different discounts. Let us denote the base discount and base sales of product $i$ predicted by the sales forecasting module as $(d_{i,t}^o, Y_{i,t}^o)$, and denote the estimated parameter of price elasticity model as $\hat{\bm{\theta}}_t$. Substituting  (\ref{non_param_predictive}) and (\ref{price_elasticity}) into 
semi-parametric model (\ref{semi_param}), we can predict the counterfactual demand as
\begin{equation}\label{counter_pred}
\ln Y_{i,t}(d_i) =  \hat{\bm{\theta}}_t^T \hat{L}_i ( \ln d_i - \ln d_{i,t}^o)+  \ln Y_{i, t}^{o}.
\end{equation}

Varying the discount from $0$ to $1$, we can predict the counterfactual sales using (\ref{counter_pred}). In the next section, we use these predicted results for optimizing discount price.
	
	\section{Multi-Period Price Optimization}
	
	\begin{figure}[tbp!] 
		\centering
		\includegraphics[width=0.9\linewidth]{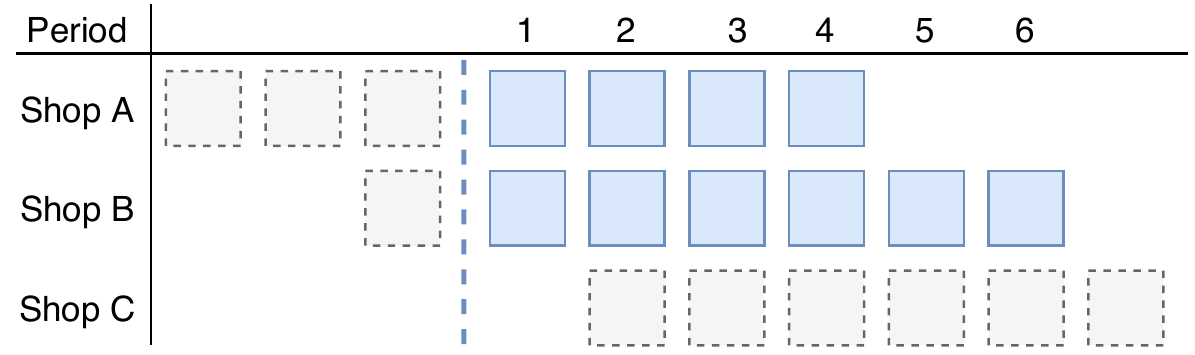} 
		\caption{An example of the activated shops and index of multiple periods. The dotted blue vertical line represents the current time. The dotted grey blocks represent the inactivate days, and solid blue block represent the activate days.} 
		\label{fig:shops} 
	\end{figure}
	
Once obtaining counterfactual demands, we can formulate a price optimization problem to maximize the overall profit. We neglect the substitution effect and assume that different products are independent. Thus, we can consider price optimization for only one product and omit the subscript $i$ for notational simplicity.

Suppose there are $|\mathcal{J}|$ shops in a region. Each shop $j \in \mathcal{J}$ 
has $B_j$ items of product A in stock and they want to sold out in $T_j$ days. If it is not sold out, it will be thrown away for ensuring the freshness of goods and leads to the economic loss. We treat one day as a period, and the number of total periods  is $T_{\text{max}} :=\arg \max_j{T_j}$. In different periods, product A has possibly different demands. Taking the targets of all shops into consideration, we aim to optimize the price  jointly and dynamically. An example of the activated shops and the index of periods is illustrated in \figurename{ \ref{fig:shops}. 
	
Let us denote the retail price of product A as  $p_0$. It is the price sold in the normal channel and is fixed. Let us denote $d_t$ as the percentage discount at period $t$, and the corresponding discount price is $p_t = p_0 d_t$.  We use the  sales forecasting algorithm presented in the  subsection  \ref{sales_forecast} to predict the sales in normal channel for all shops and all periods,  denoted as $Z_{jt}, \forall j, t$.  We use the semi-parameteric model (\ref{counter_pred}) to predict the counterfactual sales in markdown channel in period $j$ with the discount $d_t$, denoted as $Y_{jt}(d_t)$.

Since minimizing the waste loss is also a target of the retailer, we also take it into consideration. Together with the target of maximizing the overall profit, we formulate the following price optimization problem:
\begin{equation}\label{opt_problem}
\begin{split}
\max_{d_1, \dots, d_T}  & \sum_{j \in \mathcal{J}}  (\sum_{t=1}^{T_j}   p_0 d_t Y_{jt}(d_t)  -   w_j  [B_j - \sum_{t=1}^{T_j} (Y_{jt}(d_t) + Z_{jt})]^+) \\
\mbox{s.t.} & \sum_{t=1}^{T_j} (Y_{jt}(d_t) + Z_{jt}) \leq B_j, j \in \mathcal{J} \\
& lb_{jt} \leq p_t \leq ub_{jt},  t=1,\dots, T_j, j \in \mathcal{J},
\end{split}
\end{equation}
where $w_j>0$ is the weight of waste loss and $[\cdot]^+$ is the non-negative operator. The variables $lb_{jt}$ and $ub_{jt}$ are respectively minimum and maximum value of discount for shop $j$ at period $t$. 
Since both $Z_{jt}$ and $B_j$ are independent with decision variables $\{d_1, \dots , d_T\}$, we can simplify 
(\ref{opt_problem}) as 
\begin{equation}\label{simple_opt}
\begin{split}
\max_{d_1, d_2, \dots, d_T}  & \sum_{j \in \mathcal{J}}  \sum_{t=1}^{T_j}   ( p_0 d_t  + w_j) Y_{jt}(d_t)  \\
\mbox{s.t.} &\sum_{t=1}^{T_j}  (Y_{jt}(d_t) + Z_{jt}) \leq B_j, j \in \mathcal{J}  \\
& lb_{jt} \leq d_t \leq ub_{jt},  t=1,\dots, T_j, j \in \mathcal{J},
\end{split}
\end{equation}
where the first constraint describles the interaction among decision variables  $\{d_1, d_2, \dotsm , d_T\}$.

\subsection{MDP  Model}
The continuous optimization w.r.t discount is indeed not convex. However, in this paper, we do not optimize the discount directly. Considering that the candidates of the discount is finite, we transform it into discrete optimization problem instead. 
Moreover, note that the learned price-sales function $Y_{jt}(d_t)$ and the predicted normal sales $Z_{jt}$ inevitably have the randomness due to the variance and bias occurred in the learning and prediction process. In other words, the parameters in the optimization problem (\ref{simple_opt}) have uncertainty. Therefore, we use Markov decision process (MDP) to model this decision-making problem.

Without loss of generality, we assume the actual sales of shop $j$ in markdown channel and normal channel are respectively $a^y_{jt}$ and $a^z_{jt}$ at period $t$. 
Let us define the state as the inventory in stock, i.e., the state
$s_{j,t}$ represents the inventory of product A at shop $j$ at the beginning of period $t$. The initial state of shop $j$ is equal to the target sales or equivalently initial inventory $B_j$. The state $s_{j, t+1}$ is equal to the state $s_{j,t}$ subtracting both sales in the markdown channel and normal channel at period $t$, and the state $s_{j,T}$ will not be negative at  the last period. Mathematically, 
\begin{equation} \label{state_relationship}
\begin{split}
s_{j,1} & = B_j, \\
s_{j, t+1} & = s_{j, t} - a^y_{jt} - a^z_{jt}, \ t <T_j,\\
s_{j, t+1}  &= 0, \ T_j \leq t \leq T_{\text{max}}.
\end{split}
\end{equation}
Note that the value of state is non-increasing, i.e., $s_{j,1}\geq \cdots \geq s_{j,T_{j}+1}= 0$.  We define a set of finite candidates of the percentage discount  as $\mathcal{D}=\{d^1, d^2, \dots, d^M \}$. This discount set is the set of actions that the decision maker chooses from.

To modeling the  \emph{uncertainty} of sales forecasting algorithm, we dig into the historical data of the predicted sales and actual sales, and find that the sales distribution mostly follows \emph{Poisson} distribution.
In detail,  we assume the sales distribution of the markdown channel 
follows \emph{Poisson} distribution with the  parameter $Y_{jt}(d_t)$, and sales distribution
of the normal channel follows \emph{Poisson} distribution with parameter $Z_{jt}$.
Since the expectation of the Poisson distribution is equal to its parameter, we have implicitly assumed that $Y_{jt}(d_t)$ and $Z_{jt}$ are unbiased estimation of $a^y_{jt}$ and $a^z_{jt}$, respectively.

Let us define $a_{jt} := a^y_{jt} + a^z_{jt}$ as the total sales of product A in both channels.
Since  $a_{jt}$ can not be greater than inventory $s_{j,t}$, we have
\begin{equation} \label{sales_distrib}
P(a_{jt} ) = 
\left\{
\begin{array}{ll}
Poi(a_{jt}|\lambda_{jt}),  & 0 \leq a_{jt}  < s_{j,t} \\
1-\mathcal{Q}(s_{j,t}-1, \lambda_{jt}),  & a_{jt}  = s_{j,t},
\end{array}
\right.
\end{equation}
where $\lambda_{jt}:= Y_{jt}(d_t) + Z_{jt}$, $Poi(\cdot)$ is Poisson distribution and $\mathcal{Q}(\cdot, \cdot)$ is the regularized Gamma function.

Using (\ref{state_relationship}) and (\ref{sales_distrib}), we can obtain the \emph{state-transition probability} that action $d_t$ in state $s_{j,t}$  will lead to $s_{j,t+1}$:
\begin{equation}  \label{state_trans_prob}
\begin{split}
& P(s_{j,t+1} | s_{j,t}, d_{t}) \\
& = 
\left\{
\begin{array}{ll}
Poi(s_{j,t} - s_{j,t+1} |Y_{jt}(d_t) + Z_{jt} ) & 0<s_{j,t+1} \leq s_{j,t}\\
1-\mathcal{Q}(s_{j,t}-1,Y_{jt}(d_t) + Z_{jt})  & s_{j, t+1}=0.
\end{array}
\right.
\end{split}
\end{equation}
The \emph{expected immediate reward} received after transitioning from state $s_{j,t}$ to state $s_{j,t+1}$ due to action $d_t$ is 
\begin{equation} \label{reward_func}
\begin{split}
R(s_{j,t}, d_t, s_{j,t+1}) =(p_0d_t + w_j)[s_{j,t} - s_{j,t+1} - Z_{jt}]^{+}.
\end{split}
\end{equation}
The optimization problem (\ref{simple_opt})  now becomes to choose a policy $\pi(\cdot)$ that will maximize the total rewards,
\begin{equation}
\sum_{j \in \mathcal{J}} \sum_{t=1}^{T_j} R(s_{j,t},d_t, s_{j,t+1}), \ \ d_t = \pi(s_t),
\end{equation}
where  $s_t := \{s_{1,t}, \dots, s_{J,t}\}$, and $d_t$ is the action given by policy $\pi(s_t)$. To illustrate this process vividly, we give an example of Markov decision process with four states in \figurename{ \ref{fig:mdp}.
	
	\begin{figure}[tbp!] 
		\centering
		\includegraphics[width=0.9\linewidth]{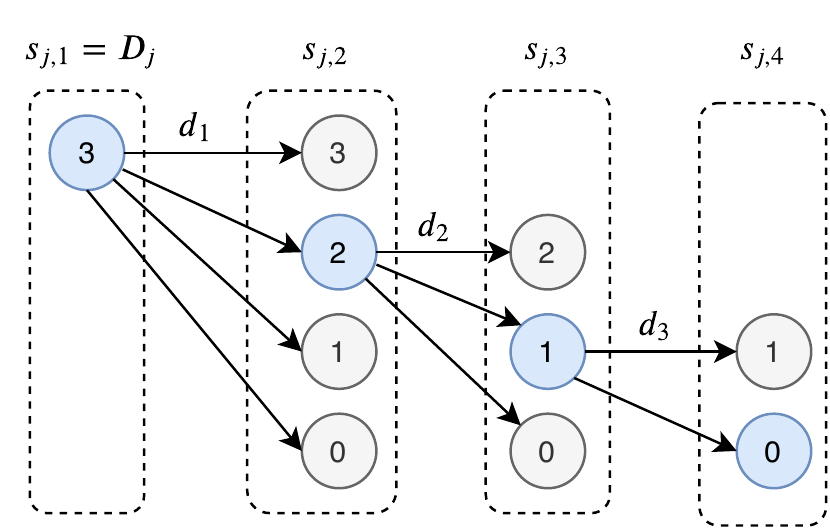} 
		\caption{An example of the Markov decision process of shop $j$ with $4$ states. Circles represent the states and the arrows represent the possible transition directions with action $d_{t}$.
			For simplicity, we only draw the arrows of blue circles.} 
		\label{fig:mdp} 
	\end{figure}

\subsection{Two-stage Algorithm}
Given state transition probability $P(\cdot)$  in (\ref{state_trans_prob}) and reward function $R(\cdot)$ in (\ref{reward_func}) for an MDP, we can seek the optimal policy through dynamic programming with Bellman equation. However, since the dimension of the state space grows exponentially as the increase of the number of the shops, a primitive backward induction method is very time-consuming.  We propose a  two-stage optimization algorithm to simplify dynamic programming, consisting of  a separate backward induction and a joint optimization.
		
\subsubsection{Separate backward induction} 
The $Q$ function of each shop $j$ is updated separately using greedy policy by backward induction from period $T_j$ to period $2$, i.e.,
\begin{equation} \label{sep_backward}
\begin{split}
Q(s_{j, t}, d_t)  & = \sum_{ s_{j, t+1} = 0}^{ s_{j,t}} P(s_{j,t+1} | s_{j,t}, d_t) (R(s_{j,t}, d_t, s_{j, t+1})  \\
& \quad +  \max_{d_{t+1} \in \mathcal{D}}Q(s_{j,t+1},d_{t+1})),   t=T_j,\dots, 2, j \in \mathcal{J},
\end{split}
\end{equation}
where the initial value is $Q(s_{j,T_{j}+1}, \cdot) = 0,  s_{j,T_{j}}=0, \forall j$. This is the bellman equation for each single shop, and it can be updated by dynamic programming efficiently by finding substructure of computation.

\subsubsection{Joint optimization}
We update the $Q$ function using the backward induction separately for each shop until the last step. In the last step (i.e., at period $1$), we can \emph{jointly} optimize the SKU price since there is only one state, i.e., $s_1=\{B_1,B_2,\cdots, B_J\}$.  The joint optimization equation can be written as follows, 
\begin{equation} \label{last_opt}
d_1^*  :=  \arg \max_{d_t \in \mathcal{D}}   \sum_{j\in \mathcal{J}}Q(s_{j,1}=B_j, d_1), 
\end{equation}
where
\begin{equation} \label{joint_opt}
\begin{split}
Q(s_{j,1}=B_j, d_1) & =  \sum_{ s_{j, 2} = 0}^{ s_{j,1}} P(s_{j,2} | s_{j,1}, d_1)  \\
& \quad \cdot (R(s_{j,1}, d_1, s_{j, 2}) d +  \max_{d_2 \in \mathcal{D}}Q(s_{j,2},d_2)).
\end{split}
\end{equation}

Thus, we obtain the optimal price $p_1^*= p_0 d_1^*$ and the expected total reward $V(s_{1})= \sum_{j \in \mathcal{J}} Q(B_j, d_1^*)$.
Since the pricing algorithm is performed daily as soon as the new data arrived for each new day, it will recompute the optimal price for the first period which is satisfied the price constraint, i.e., the discount price of all shops are equal. Therefore, it is not necessary that the obtained prices of different shops are not equal from period 2 to the last period.

For clarity, the pricing algorithm for markdowns in fresh retail is summarized in Algorithm \ref{alg_pricing}, where we refill the subscript $i$ of all variables to represent  product $i$ for completeness.

\subsection{Time Complexity}
In this subsection, we analyze the time complexity of the proposed two-stage algorithm. There are at most $B_j^2 * M$  iterations to compute $\mathcal{Q}(s_{t},d_t)$, where $M$ is the number of actions. So the time complexity of two-stage algorithm is $O(T_{\text{max}} * M * B_{\text{max}}^2 * |\mathcal{J}| )$, where $B_{\text{max}} := \arg \max_j B_j $. In comparison, note that the state space for a primitive backward induction method is vector space, and the number of state is $\prod_{j \in \mathcal{J}} B_j$. The time complexity for this method is  $O(T_{\text{max}} * M * {B_{\text{max}}}^{2|\mathcal{J}|} )$, which grows exponentially with the increase of the number of the shops.  
It is easy to see that our algorithm is much more efficient than the traditional dynamic programming algorithm when $|\mathcal{J}|$ is large. In practice, our algorithm can be further pruned by reducing redundant computation.

\subsection{Extension}
The \emph{counterfactual prediction and  optimization} framework is quite general and can be applied into other scenarios of e-commerce platform:
\begin{itemize}
\item In fresh retail scenario, this approach can also be applied for the markdown of daily-fresh goods, which has only one day shelf life. It aims to boost sales by cutting the price before the end of the day, such as only two hours left.

\item In the marketing scenario, it can be applied for personalized coupon assignment task, which aims to  maximize the overall Return on Investment (ROI) of the platform by assigning the different coupons to different users with budget constraint.

\item In the  customer service scenario, it can be applied for personalized compensation pay-outs task, which aims to maximize the satisfaction rate of customers by offering the optimal pay-outs for users with the budget constraint.

\end{itemize}
In above three examples, the treatment is price, coupon and compensation pay-outs respectively,
the outcome is the sales of product, the conversion rate and satisfaction rate  of the users respectively, and the optimization objective is overall profit, ROI and satisfaction rate respectively. 
To solve these problems,  we first build the causal relationship between treatment and outcome based on semi-parametric model, and then optimize the objective with budget or inventory constraints.  It is worth noting that  we have presented a marketing budget allocation framework for the second scenario  in our previous work \cite{zhao2019unified}. While, in this paper, the proposed counterfactual prediction-based framework is more general and can be applied into more scenarios.

\begin{algorithm}[tb]
\caption{Pricing algorithm for markdowns of e-commerce retails} \label{alg_pricing} 
\begin{algorithmic}[1] %[1] enables line numbers
	\Require{ $\lambda, \tau, \omega, \bm{x}_{i, \cdot}, Y_{i, \cdot}^{obs}, Y_{i, \cdot}^{nor}, d_{i, \cdot}^{o}, \forall i$.}
	\Ensure{$p_{i, t}^*, \forall i, \forall t.$}
	\Statex   
	\State Pretrain $h(\cdot)$ and $g(\cdot)$ using historical observational  data.
	\For {$t=1,2, \dots $}
	\State Receive new data $\bm{x}_{i,t}, Y_{i,t}^{obs}, Y_{i,t}^{nor}, d_{i,t}^{o}, \forall i$.
	\State  Update $ h(\cdot)$ and predict $(d_{i,t}^o, Y_{i,t}^o)$ using (\ref{non_param_predictive}).
	\State   Update  $\hat{\bm{\theta}}_t$ using (\ref{rls_objective}).
	\For{$i=1,\dots, N$}
	\For{$k=1,2,\dots, T_{i, \text{max}}-t+1$}
	\State Predict $Z_{ijk}$ and $Y_{ijk}(d_{ik}), \forall d_{ik} \in \mathcal{D}_i$.
	\EndFor
	\State Optimize $d_{i,1}^*$ using (\ref{sep_backward})  and (\ref{last_opt}).
	\EndFor
	\State $p_{i, t}^* = p_{i, 0} d_{i, 1}^*$.
	\EndFor
	\Statex 	
\end{algorithmic}
\end{algorithm}

\section{Experiment} 
To evaluate the performance of the proposed approach, we  apply the markdown pricing algorithm in a real-world e-commerce fresh retail - Freshippo. To reduce cost and stimulate consumption,  Freshippo will give a discount for the product whose sales performance is not satisfactory. It will be sold in markdown channel, where customers can buy goods by discount if their total purchase reached a certain amount.  Our aim is to help the retailer of Freshippo to decide which optimal discount price of a product to be set to maximize the overall profit. 

\subsection{Offline Experiment}
To optimize the price with business constraints, the first step and the key step is to learn the price-demand curve as described in Section \ref{sec_demand_learning}. We use the offline transaction data of Freshippo to train and evaluate the proposed semi-parametric structural demand model.  
We collect observational data in markdown channel over more than $100$ fresh stores across $10$ different cities in China over $6$ months.  There are about $11000$+ SKUs in total (we omit the exact number for privacy-preserving).  The detail of feature extraction is provided in Supplement.

\paragraph{Learning model} Using the extracted features, we use XGBoost \cite{chen2016xgboost} as our base sales forecasting model. The deep neural network is also recommended when there are sufficient data samples  for training \cite{bengio2013representation, goodfellow2016deep}. Using the predicted base-discount and sales as inputs, we solve the objective (\ref{rls_objective}) by recursive least squares algorithm, and the price elasticity $\bm{\theta}$ is updated in the recursion fashion.  In the Freshippo scenario, the price elasticity is daily updated once the new transaction data is collected and  features are automatically extracted.

\paragraph{Evaluation metric}
To evaluate the predictive ability of our semi-parametric model,
we predict the sales of the next day in markdown channel with the actual discount of that day. 
The results are measured in Relative Mean Absolute Error (RMAE), defined as:
\begin{equation}
\mbox{RMAE} = \frac{\sum_{i=1}^N  |Y_{i} - \hat{Y}_{i}|}{\sum_{i=1}^N Y_{i}},
\end{equation}
where $N$ is the number of the test samples, $Y_i$ is the ground truth sales and $\hat{Y}_i$ is the predicted sales. 

\paragraph{Parameter settings}
To tun the model hyperparameters, we split data into training, validation and test data according to the time: first $65\%$ for training, next $15\%$ for validation and last $20\%$ for testing. For price elasticity model, we empirically set the forgetting factor $\tau=0.95$, the regularization parameter $\lambda=0.5$.

\paragraph{Comparison algorithm.}  We compare the proposed model with the classical boost tree model and deep model.
\begin{itemize}
	\item XGBoost. This powerful boost tree model has been applied in many industrial applications. For counterfactual prediction task, we treat the discount variable as one of its input features,  and predict the outcome by varying discount with other features fixed.   For the sake of fairness,  its hyperparameters are set the same as our base sales forecasting model. 
	\item DeepIV \cite{hartford2017deep}. This method use deep model to characterize the relationships between treatment and outcome in the presence of instrument variables. We use the average price of third-level category as the instrument variables, and use Gaussian distribution for continuous treatment.
\end{itemize}

\begin{figure}[tbp!] 
	\centering   
			\includegraphics[width=0.9\linewidth]{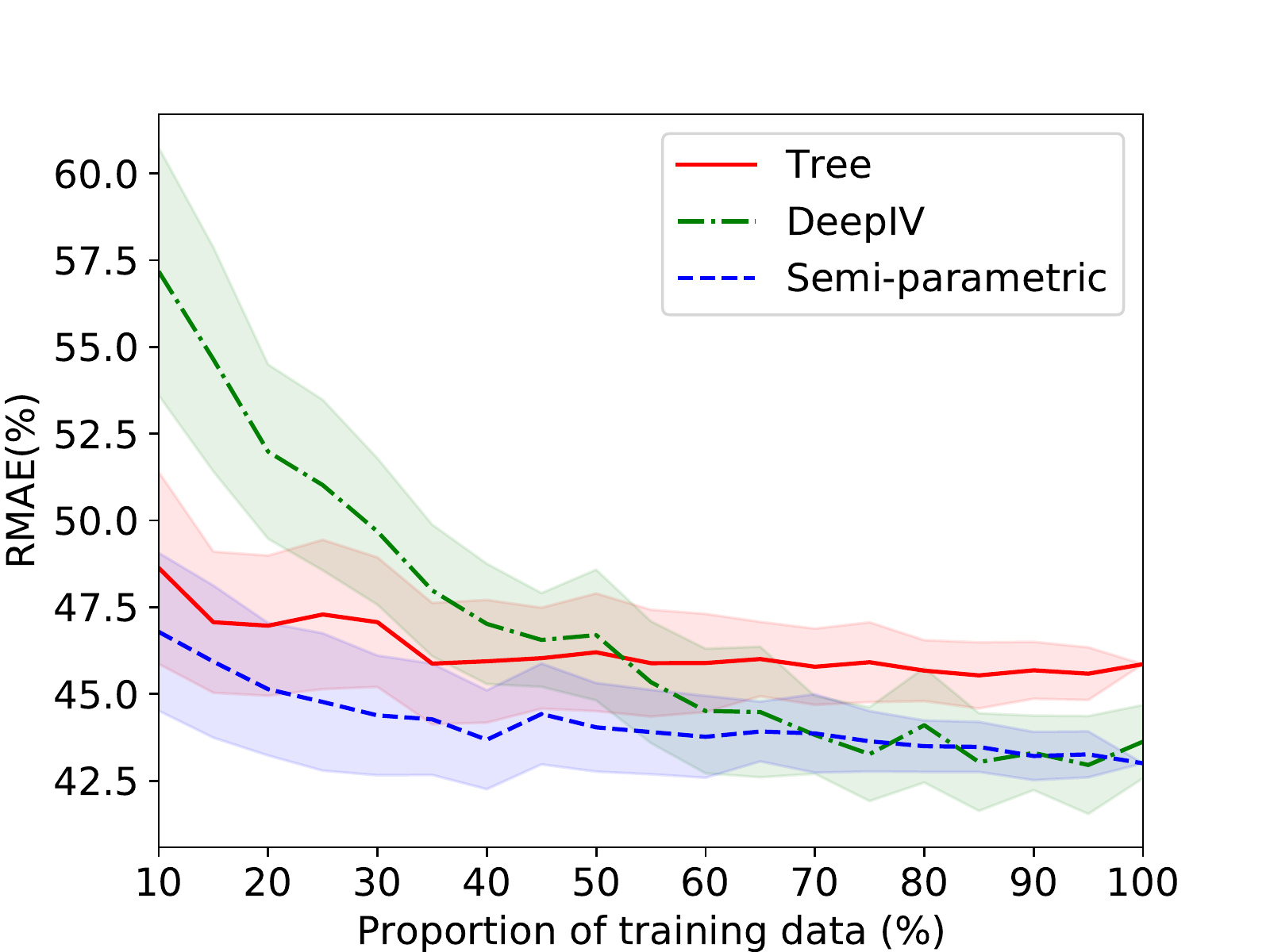}
	\caption{The comparisons between tree model,  deep model,  and our semi-parametric model under different proportion of training data.} 		
	\label{fig:pred_err}
\end{figure}

\paragraph{\textbf{Results:} }
We first evaluate the prediction error of each algorithm with different proportion of training data.  
As shown in  \figurename{ \ref{fig:pred_err}}, the deep model has poor performance with small training data. The tree model is relatively insensitive with the data size but its performance is poor even when there are more data available. Our proposed semi-parametric model achieve the best RMAE performance and its variance is reduced with the increase of the number of training data. To further illustrate the interpretability of the proposed model, we plot four price-sales curves of randomly chosen SKUs in a random store in \figurename{ \ref{fig:curve}.  The curve learned by the tree model has unpredictable jitter, and its  outcome keeps unchanged in a large range of discount (0.5-0.6, 0.8-1.0). The curve learned by deepIV is almost a line, which  indicates the sales is independent with the price. Both tree and deep model can not correctly reveal the price-sales relationship and the inferred results are not credible. While, our semi-parametric model is much smoother and reveals a monotonous relationship between price and sales. The variation between close prices is small, which is consistent with the intuition.

\begin{figure*}[tbp!] 
	\centering   
	\subfigure[Tree model]
	{
		\label{fig:xgb_curve}
		\includegraphics[width=0.3\linewidth]{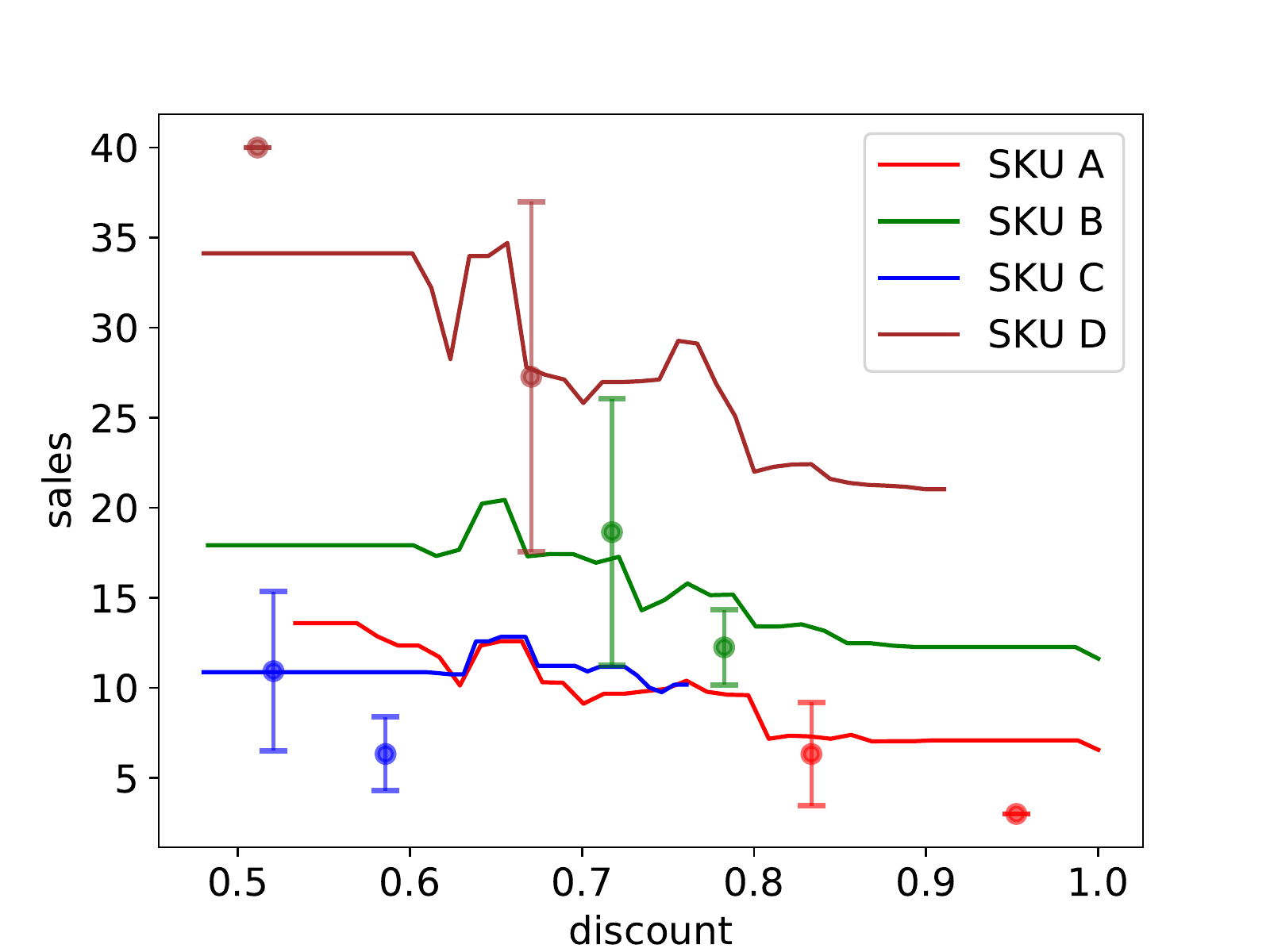}
	}
	\subfigure[DeepIV model]
{
	\label{fig:deepiv_curve}
	\includegraphics[width=0.3\linewidth]{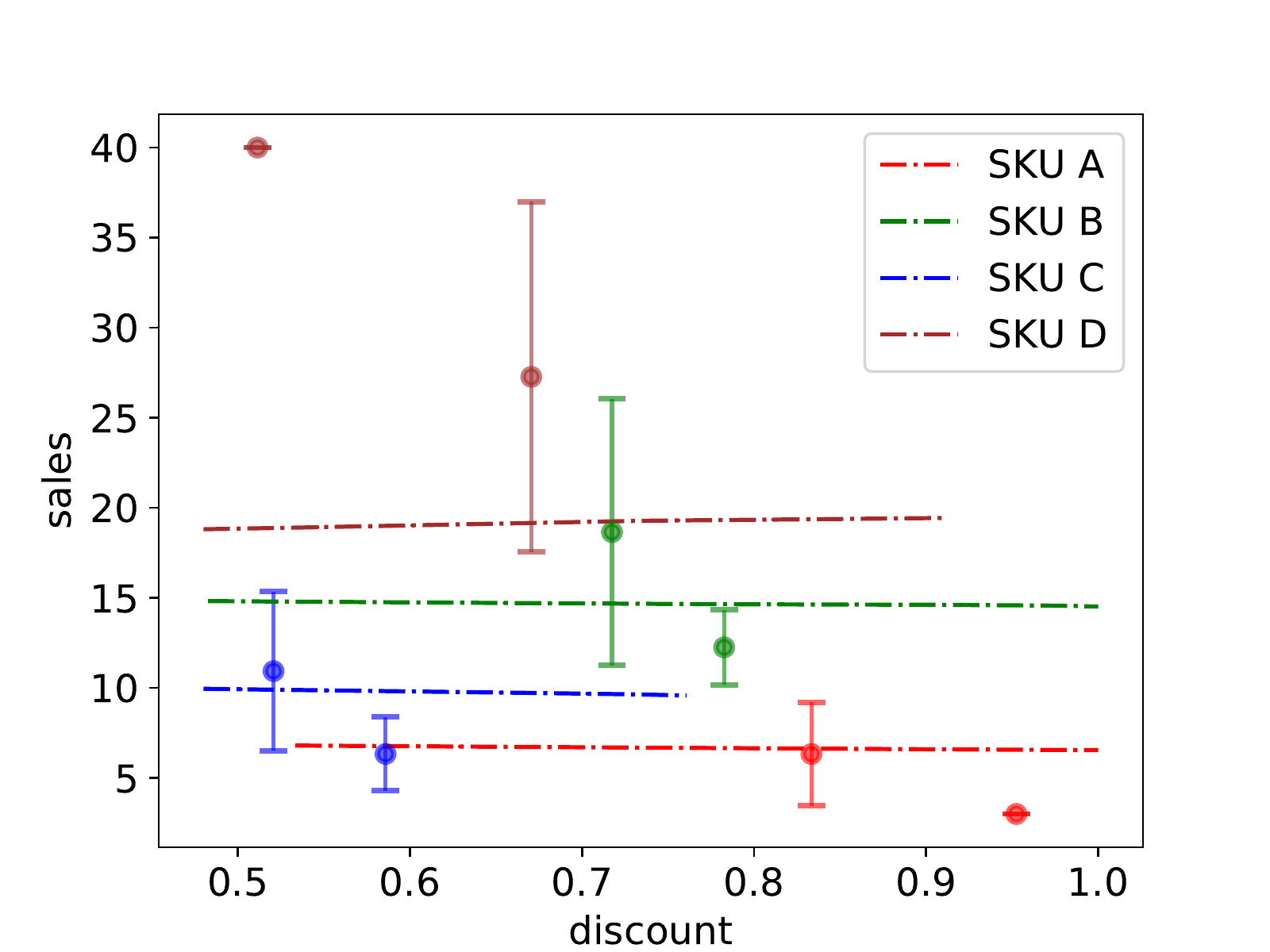}
}
	\subfigure[Proposed semi-parametric model]
	{
		\label{fig:linear_curve}
		\includegraphics[width=0.3\linewidth]{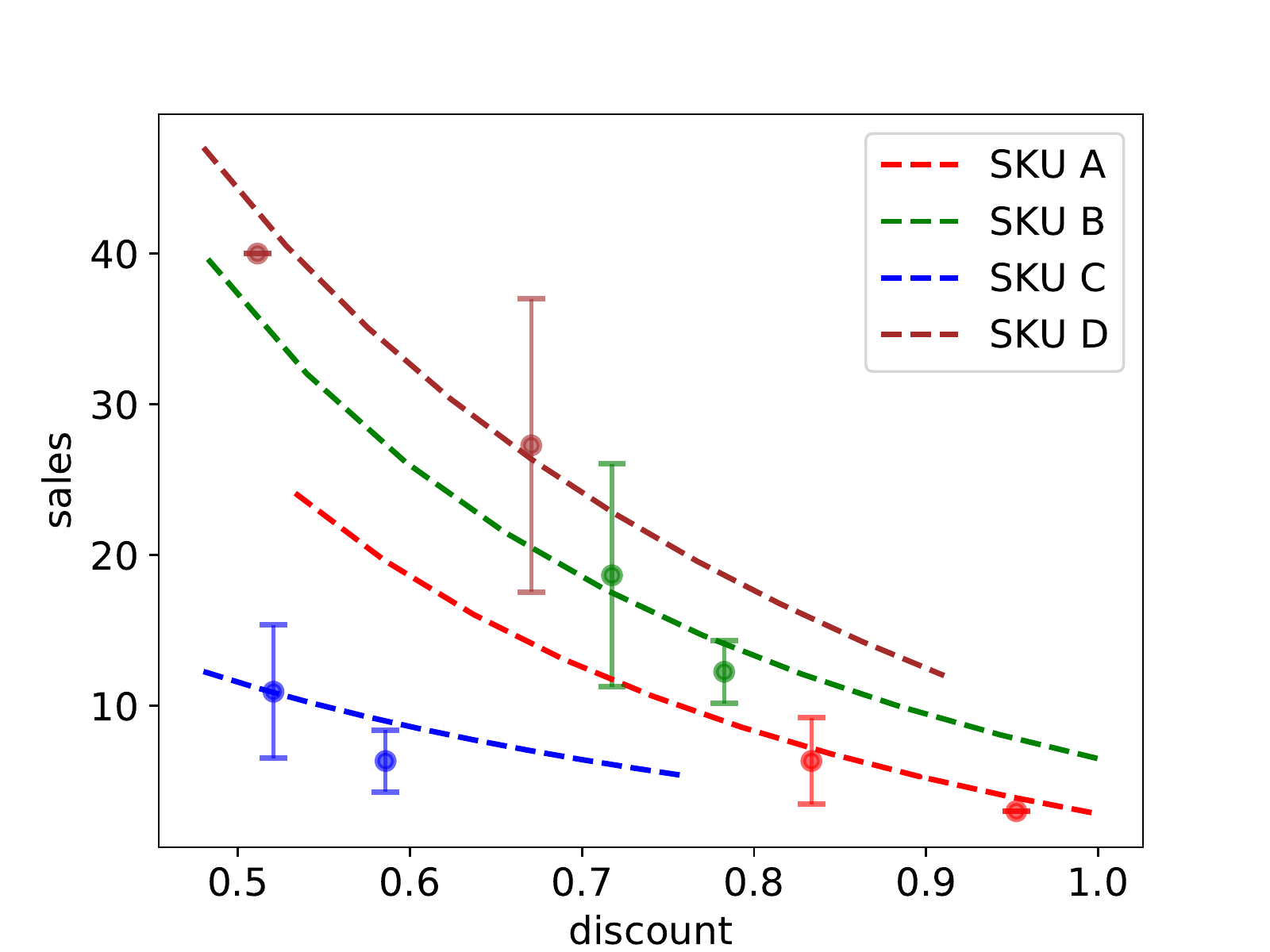}
	}
	\caption{Price-sales curves of four randomly chosen SKUs. The error bar denotes the mean and the standard deviation of historical observed sales.} 
	\label{fig:curve}
\end{figure*}

\subsection{Online A/B Testing}
We evaluate our markdown approach in a online fresh retail, Freshippo, as shown in \figurename{ \ref{fig:hema}. Traditionally, the retailer of Freshippo uses a manual pricing strategy that chooses an empirical discount, such as $30\% $ off or $50\%$ off. 
Although the manual pricing strategy is simple, it is the fruit of experience of operations for many years and it works in most cases. The target products, which are required for markdowns, are provided by the inventory control system of Freshippo. The target sales and min-max price constraints are given by fresh operational staff.  Once got the signal of markdowns, the intelligence marketing system of Freshippo will call our pricing algorithm to offer the optimal discount.  Our algorithm has been applied in the fresh stores in four regions, Beijing, Shanghai, Hangzhou and Shenzhou.  
	
For fair comparison, we design the online A/B testing. The products required to be taken a discount are randomly assigned to the operation's manual approach and our markdown approach.

\paragraph{Evaluation metrics}
To evaulate the performance of the proposed algorithm,  we use the target completion rate (TCR) as  the metric, defined as
\begin{equation}
\begin{split}
\mbox{TCR}_{nor} &= \frac{\sum_{i=1}^N \mbox{\#SALES}_{i, nor}}{\sum_{j=1}^N B_i}, \\
\mbox{TCR}_{md} &= \frac{\sum_{i=1}^N \mbox{\#SALES}_{i, md}}{\sum_{j=1}^N B_i}, 
\end{split}
\end{equation}
where $\mbox{\#SALES}_{i, nor}$ and $\mbox{\#SALES}_{i, md}$  denote the sales in the normal channel and the sales in markdown channel, respectively. 
The target completion rate assesses the effect on clearance of markdowns.
Besides, we also evaluate the ratio of Gross Merchandise Volume (GMV) between normal channel and makdown channel, namely
\begin{equation}
\mbox{GMV\_IMP} = \frac{\sum_{i=1}^N  d_i p_i  \mbox{\#SALES}_{i, md}}{\sum_{i=1}^N p_i \mbox{\#SALES}_{i, nor}}.
\end{equation}
This metric evaluates the GMV improvement after markdowns.

\paragraph{ \textbf{Results} }	

\begin{table}[!t]
	\centering
	\caption{Target completion rate and GMV improvement} 
	\label{tab_gmv_imp}
	\begin{tabular}{c|c|c|c|c}
		\hline
%				& \multicolumn{3}{c|}{Target completion rate} & GMV improvement   \\ \hline 
		 & $\mbox{TCR}_{nor}$ & $\mbox{TCR}_{md}$  & $\mbox{TCR}_{total}$ & GMV\_IMP  \\ \hline
		Operations  &  34.25\%&  46.68\% &  80.93\% &  101.49\% \\ 
		Ours &  38.92\% &   53.20\% & \textbf{92.12\%} &  \textbf{118.63\%} \\ \hline
	\end{tabular}
\end{table}

The A/B testing is carried out about $2$ months.  As shown in \tablename{\ref{tab_gmv_imp}, 
 the target completion rate of operation's group is about $34.25\%$ before markdown, and this number goes to $80.93\%$ after markdowns. While the target completion rate of our group is about $38.92\%$  before markdown, and this number goes to  $92.12\%$ after markdown. 
It indicates that our approach can better achieve the clearance target than the maunal approach. Besides,  the GMV improvement of our pricing algorithm  is higher about $17.14\%$ than that of maunal approach.  This result shows that our approach can help retailer obtain the higher GMV/profit. 

In summary, our pricing algorithm is easy to explan as it has interpretable price-sales curve, and it is intelligent as it can automatically offer optimal price with high profit in consideration of complex business constraints, such as inventory constraint.

\section{Conclusion}
In this paper, we present a novel pricing framework for markdowns in e-commerce
fresh retails, consisting of counterfactual prediction and multi-period price optimization.  Firstly, we propose a data-driven semi-parametric structural demand model, which has both predictability and interpretability.  The proposed demand model reveals the relationship between price and sales and can be used for predicting counterfactual demand with different prices. 
The proposed counterfactual model has a lower model complexity and a clear causal structure, therefore it is much more interpretable than traditional ML and deep model.
Secondly, we take the demand uncertainty into consideration, and present a dynamic pricing strategy of perishable products in a multi-period setting. An efficient two-stage algorithm is proposed for solving the MDP problem. It reduces the time complexity from exponential to polynomial. Finally, we apply our framework into a real-world e-commerce fresh retail. Both offline experiment and online A/B testing show the effectiveness of our approach.

%\section{Acknowledgments}
%xxxxx

\bibliographystyle{ACM-Reference-Format}
\bibliography{mybibfile}


\begin{thebibliography}{37}

%%% the Web address.

\ifx \showCODEN    \undefined \def \showCODEN     #1{\unskip}     \fi
\ifx \showDOI      \undefined \def \showDOI       #1{#1}\fi
\ifx \showISBNx    \undefined \def \showISBNx     #1{\unskip}     \fi
\ifx \showISBNxiii \undefined \def \showISBNxiii  #1{\unskip}     \fi
\ifx \showISSN     \undefined \def \showISSN      #1{\unskip}     \fi
\ifx \showLCCN     \undefined \def \showLCCN      #1{\unskip}     \fi
\ifx \shownote     \undefined \def \shownote      #1{#1}          \fi
\ifx \showarticletitle \undefined \def \showarticletitle #1{#1}   \fi
\ifx \showURL      \undefined \def \showURL       {\relax}        \fi
% invisible to TeX
\providecommand\bibfield[2]{#2}
\providecommand\bibinfo[2]{#2}
\providecommand\natexlab[1]{#1}
\providecommand\showeprint[2][]{arXiv:#2}

\bibitem[\protect\citeauthoryear{Athey and Imbens}{Athey and Imbens}{2016}]%
        {athey2016recursive}
\bibfield{author}{\bibinfo{person}{Susan Athey} {and} \bibinfo{person}{Guido
  Imbens}.} \bibinfo{year}{2016}\natexlab{}.
\newblock \showarticletitle{Recursive partitioning for heterogeneous causal
  effects}.
\newblock \bibinfo{journal}{\emph{Proceedings of the National Academy of
  Sciences}} \bibinfo{volume}{113}, \bibinfo{number}{27}
  (\bibinfo{year}{2016}), \bibinfo{pages}{7353--7360}.
\newblock


\bibitem[\protect\citeauthoryear{Athey, Tibshirani, Wager, et~al\mbox{.}}{Athey
  et~al\mbox{.}}{2019}]%
        {athey2019generalized}
\bibfield{author}{\bibinfo{person}{Susan Athey}, \bibinfo{person}{Julie
  Tibshirani}, \bibinfo{person}{Stefan Wager}, {et~al\mbox{.}}}
  \bibinfo{year}{2019}\natexlab{}.
\newblock \showarticletitle{Generalized random forests}.
\newblock \bibinfo{journal}{\emph{The Annals of Statistics}}
  \bibinfo{volume}{47}, \bibinfo{number}{2} (\bibinfo{year}{2019}),
  \bibinfo{pages}{1148--1178}.
\newblock


\bibitem[\protect\citeauthoryear{Bengio, Courville, and Vincent}{Bengio
  et~al\mbox{.}}{2013}]%
        {bengio2013representation}
\bibfield{author}{\bibinfo{person}{Yoshua Bengio}, \bibinfo{person}{Aaron
  Courville}, {and} \bibinfo{person}{Pascal Vincent}.}
  \bibinfo{year}{2013}\natexlab{}.
\newblock \showarticletitle{Representation learning: A review and new
  perspectives}.
\newblock \bibinfo{journal}{\emph{IEEE transactions on pattern analysis and
  machine intelligence}} \bibinfo{volume}{35}, \bibinfo{number}{8}
  (\bibinfo{year}{2013}), \bibinfo{pages}{1798--1828}.
\newblock


\bibitem[\protect\citeauthoryear{Blattberg and George}{Blattberg and
  George}{2010}]%
        {blattberg2010shrinkage}
\bibfield{author}{\bibinfo{person}{Robert~C Blattberg} {and}
  \bibinfo{person}{Edward~I George}.} \bibinfo{year}{2010}\natexlab{}.
\newblock \showarticletitle{Shrinkage estimation of price and promotional
  elasticities: Seemingly unrelated equations}.
\newblock In \bibinfo{booktitle}{\emph{Perspectives On Promotion And Database
  Marketing: The Collected Works of Robert C Blattberg}}.
  \bibinfo{publisher}{World Scientific}, \bibinfo{pages}{145--156}.
\newblock


\bibitem[\protect\citeauthoryear{Bolton and Shankar}{Bolton and
  Shankar}{2003}]%
        {bolton2003empirically}
\bibfield{author}{\bibinfo{person}{Ruth~N Bolton} {and}
  \bibinfo{person}{Venkatesh Shankar}.} \bibinfo{year}{2003}\natexlab{}.
\newblock \showarticletitle{An empirically derived taxonomy of retailer pricing
  and promotion strategies}.
\newblock \bibinfo{journal}{\emph{Journal of Retailing}} \bibinfo{volume}{79},
  \bibinfo{number}{4} (\bibinfo{year}{2003}), \bibinfo{pages}{213--224}.
\newblock


\bibitem[\protect\citeauthoryear{Caro and Gallien}{Caro and Gallien}{2012}]%
        {caro2012clearance}
\bibfield{author}{\bibinfo{person}{Felipe Caro} {and}
  \bibinfo{person}{J{\'e}r{\'e}mie Gallien}.} \bibinfo{year}{2012}\natexlab{}.
\newblock \showarticletitle{Clearance pricing optimization for a fast-fashion
  retailer}.
\newblock \bibinfo{journal}{\emph{Operations Research}} \bibinfo{volume}{60},
  \bibinfo{number}{6} (\bibinfo{year}{2012}), \bibinfo{pages}{1404--1422}.
\newblock


\bibitem[\protect\citeauthoryear{Chen and Guestrin}{Chen and Guestrin}{2016}]%
        {chen2016xgboost}
\bibfield{author}{\bibinfo{person}{Tianqi Chen} {and} \bibinfo{person}{Carlos
  Guestrin}.} \bibinfo{year}{2016}\natexlab{}.
\newblock \showarticletitle{Xgboost: A scalable tree boosting system}. In
  \bibinfo{booktitle}{\emph{Proceedings of the 22nd acm sigkdd international
  conference on knowledge discovery and data mining}}. ACM,
  \bibinfo{pages}{785--794}.
\newblock


\bibitem[\protect\citeauthoryear{Elmachtoub and Grigas}{Elmachtoub and
  Grigas}{2017}]%
        {elmachtoub2017smart}
\bibfield{author}{\bibinfo{person}{Adam~N Elmachtoub} {and}
  \bibinfo{person}{Paul Grigas}.} \bibinfo{year}{2017}\natexlab{}.
\newblock \showarticletitle{Smart" predict, then optimize"}.
\newblock \bibinfo{journal}{\emph{arXiv preprint arXiv:1710.08005}}
  (\bibinfo{year}{2017}).
\newblock


\bibitem[\protect\citeauthoryear{Fan, Che, and Chen}{Fan et~al\mbox{.}}{2017}]%
        {fan2017product}
\bibfield{author}{\bibinfo{person}{Zhi-Ping Fan}, \bibinfo{person}{Yu-Jie Che},
  {and} \bibinfo{person}{Zhen-Yu Chen}.} \bibinfo{year}{2017}\natexlab{}.
\newblock \showarticletitle{Product sales forecasting using online reviews and
  historical sales data: A method combining the Bass model and sentiment
  analysis}.
\newblock \bibinfo{journal}{\emph{Journal of Business Research}}
  \bibinfo{volume}{74} (\bibinfo{year}{2017}), \bibinfo{pages}{90--100}.
\newblock


\bibitem[\protect\citeauthoryear{Ferreira, Lee, and Simchi-Levi}{Ferreira
  et~al\mbox{.}}{2016}]%
        {ferreira2016analytics}
\bibfield{author}{\bibinfo{person}{Kris~Johnson Ferreira}, \bibinfo{person}{Bin
  Hong~Alex Lee}, {and} \bibinfo{person}{David Simchi-Levi}.}
  \bibinfo{year}{2016}\natexlab{}.
\newblock \showarticletitle{Analytics for an online retailer: Demand
  forecasting and price optimization}.
\newblock \bibinfo{journal}{\emph{Manufacturing \& Service Operations
  Management}} \bibinfo{volume}{18}, \bibinfo{number}{1}
  (\bibinfo{year}{2016}), \bibinfo{pages}{69--88}.
\newblock


\bibitem[\protect\citeauthoryear{Fisher, Gallino, and Li}{Fisher
  et~al\mbox{.}}{2018}]%
        {fisher2018competition}
\bibfield{author}{\bibinfo{person}{Marshall Fisher}, \bibinfo{person}{Santiago
  Gallino}, {and} \bibinfo{person}{Jun Li}.} \bibinfo{year}{2018}\natexlab{}.
\newblock \showarticletitle{Competition-based dynamic pricing in online
  retailing: A methodology validated with field experiments}.
\newblock \bibinfo{journal}{\emph{Management Science}} \bibinfo{volume}{64},
  \bibinfo{number}{6} (\bibinfo{year}{2018}), \bibinfo{pages}{2496--2514}.
\newblock


\bibitem[\protect\citeauthoryear{Goodfellow, Bengio, and Courville}{Goodfellow
  et~al\mbox{.}}{2016}]%
        {goodfellow2016deep}
\bibfield{author}{\bibinfo{person}{Ian Goodfellow}, \bibinfo{person}{Yoshua
  Bengio}, {and} \bibinfo{person}{Aaron Courville}.}
  \bibinfo{year}{2016}\natexlab{}.
\newblock \bibinfo{booktitle}{\emph{Deep learning}}.
\newblock \bibinfo{publisher}{MIT press}.
\newblock


\bibitem[\protect\citeauthoryear{Hartford, Lewis, Leyton-Brown, and
  Taddy}{Hartford et~al\mbox{.}}{2017}]%
        {hartford2017deep}
\bibfield{author}{\bibinfo{person}{Jason Hartford}, \bibinfo{person}{Greg
  Lewis}, \bibinfo{person}{Kevin Leyton-Brown}, {and} \bibinfo{person}{Matt
  Taddy}.} \bibinfo{year}{2017}\natexlab{}.
\newblock \showarticletitle{Deep IV: A flexible approach for counterfactual
  prediction}. In \bibinfo{booktitle}{\emph{Proceedings of the 34th
  International Conference on Machine Learning-Volume 70}}. JMLR. org,
  \bibinfo{pages}{1414--1423}.
\newblock


\bibitem[\protect\citeauthoryear{Hernan and Robins}{Hernan and Robins}{2010}]%
        {hernan2010causal}
\bibfield{author}{\bibinfo{person}{Miguel~A Hernan} {and}
  \bibinfo{person}{James~M Robins}.} \bibinfo{year}{2010}\natexlab{}.
\newblock \bibinfo{title}{Causal inference}.
\newblock
\newblock


\bibitem[\protect\citeauthoryear{Hofman, Sharma, and Watts}{Hofman
  et~al\mbox{.}}{2017}]%
        {hofman2017prediction}
\bibfield{author}{\bibinfo{person}{Jake~M Hofman}, \bibinfo{person}{Amit
  Sharma}, {and} \bibinfo{person}{Duncan~J Watts}.}
  \bibinfo{year}{2017}\natexlab{}.
\newblock \showarticletitle{Prediction and explanation in social systems}.
\newblock \bibinfo{journal}{\emph{Science}} \bibinfo{volume}{355},
  \bibinfo{number}{6324} (\bibinfo{year}{2017}), \bibinfo{pages}{486--488}.
\newblock


\bibitem[\protect\citeauthoryear{Hruschka}{Hruschka}{2006}]%
        {hruschka2006relevance}
\bibfield{author}{\bibinfo{person}{Harald Hruschka}.}
  \bibinfo{year}{2006}\natexlab{}.
\newblock \showarticletitle{Relevance of functional flexibility for
  heterogeneous sales response models: a comparison of parametric and
  semi-nonparametric models}.
\newblock \bibinfo{journal}{\emph{European journal of operational research}}
  \bibinfo{volume}{174}, \bibinfo{number}{2} (\bibinfo{year}{2006}),
  \bibinfo{pages}{1009--1020}.
\newblock


\bibitem[\protect\citeauthoryear{Huang, Leng, and Parlar}{Huang
  et~al\mbox{.}}{2013}]%
        {huang2013demand}
\bibfield{author}{\bibinfo{person}{Jian Huang}, \bibinfo{person}{Mingming
  Leng}, {and} \bibinfo{person}{Mahmut Parlar}.}
  \bibinfo{year}{2013}\natexlab{}.
\newblock \showarticletitle{Demand functions in decision modeling: A
  comprehensive survey and research directions}.
\newblock \bibinfo{journal}{\emph{Decision Sciences}} \bibinfo{volume}{44},
  \bibinfo{number}{3} (\bibinfo{year}{2013}), \bibinfo{pages}{557--609}.
\newblock


\bibitem[\protect\citeauthoryear{Ito and Fujimaki}{Ito and Fujimaki}{2016}]%
        {ito2016large}
\bibfield{author}{\bibinfo{person}{Shinji Ito} {and} \bibinfo{person}{Ryohei
  Fujimaki}.} \bibinfo{year}{2016}\natexlab{}.
\newblock \showarticletitle{Large-scale price optimization via network flow}.
  In \bibinfo{booktitle}{\emph{Advances in Neural Information Processing
  Systems}}. \bibinfo{pages}{3855--3863}.
\newblock


\bibitem[\protect\citeauthoryear{Ito and Fujimaki}{Ito and Fujimaki}{2017}]%
        {ito2017optimization}
\bibfield{author}{\bibinfo{person}{Shinji Ito} {and} \bibinfo{person}{Ryohei
  Fujimaki}.} \bibinfo{year}{2017}\natexlab{}.
\newblock \showarticletitle{Optimization beyond prediction: Prescriptive price
  optimization}. In \bibinfo{booktitle}{\emph{Proceedings of the 23rd ACM
  SIGKDD International Conference on Knowledge Discovery and Data Mining}}.
  ACM, \bibinfo{pages}{1833--1841}.
\newblock


\bibitem[\protect\citeauthoryear{LeCun, Bengio, and Hinton}{LeCun
  et~al\mbox{.}}{2015}]%
        {lecun2015deep}
\bibfield{author}{\bibinfo{person}{Yann LeCun}, \bibinfo{person}{Yoshua
  Bengio}, {and} \bibinfo{person}{Geoffrey Hinton}.}
  \bibinfo{year}{2015}\natexlab{}.
\newblock \showarticletitle{Deep learning}.
\newblock \bibinfo{journal}{\emph{nature}} \bibinfo{volume}{521},
  \bibinfo{number}{7553} (\bibinfo{year}{2015}), \bibinfo{pages}{436}.
\newblock


\bibitem[\protect\citeauthoryear{Li and Huh}{Li and Huh}{2011}]%
        {li2011pricing}
\bibfield{author}{\bibinfo{person}{Hongmin Li} {and}
  \bibinfo{person}{Woonghee~Tim Huh}.} \bibinfo{year}{2011}\natexlab{}.
\newblock \showarticletitle{Pricing multiple products with the multinomial
  logit and nested logit models: Concavity and implications}.
\newblock \bibinfo{journal}{\emph{Manufacturing \& Service Operations
  Management}} \bibinfo{volume}{13}, \bibinfo{number}{4}
  (\bibinfo{year}{2011}), \bibinfo{pages}{549--563}.
\newblock


\bibitem[\protect\citeauthoryear{Louizos, Shalit, Mooij, Sontag, Zemel, and
  Welling}{Louizos et~al\mbox{.}}{2017}]%
        {louizos2017causal}
\bibfield{author}{\bibinfo{person}{Christos Louizos}, \bibinfo{person}{Uri
  Shalit}, \bibinfo{person}{Joris~M Mooij}, \bibinfo{person}{David Sontag},
  \bibinfo{person}{Richard Zemel}, {and} \bibinfo{person}{Max Welling}.}
  \bibinfo{year}{2017}\natexlab{}.
\newblock \showarticletitle{Causal effect inference with deep latent-variable
  models}. In \bibinfo{booktitle}{\emph{Advances in neural information
  processing systems}}. \bibinfo{pages}{6446--6456}.
\newblock


\bibitem[\protect\citeauthoryear{Ma, Fildes, and Huang}{Ma
  et~al\mbox{.}}{2016}]%
        {ma2016demand}
\bibfield{author}{\bibinfo{person}{Shaohui Ma}, \bibinfo{person}{Robert
  Fildes}, {and} \bibinfo{person}{Tao Huang}.} \bibinfo{year}{2016}\natexlab{}.
\newblock \showarticletitle{Demand forecasting with high dimensional data: The
  case of SKU retail sales forecasting with intra-and inter-category
  promotional information}.
\newblock \bibinfo{journal}{\emph{European Journal of Operational Research}}
  \bibinfo{volume}{249}, \bibinfo{number}{1} (\bibinfo{year}{2016}),
  \bibinfo{pages}{245--257}.
\newblock


\bibitem[\protect\citeauthoryear{Mart{\'\i}nez-Ruiz, Moll{\'a}-Descals,
  G{\'o}mez-Borja, and Rojo-{\'A}lvarez}{Mart{\'\i}nez-Ruiz
  et~al\mbox{.}}{2006}]%
        {martinez2006evaluating}
\bibfield{author}{\bibinfo{person}{Mar{\'\i}a~Pilar Mart{\'\i}nez-Ruiz},
  \bibinfo{person}{Alejandro Moll{\'a}-Descals}, \bibinfo{person}{MA
  G{\'o}mez-Borja}, {and} \bibinfo{person}{Jos{\'e}~Luis Rojo-{\'A}lvarez}.}
  \bibinfo{year}{2006}\natexlab{}.
\newblock \showarticletitle{Evaluating temporary retail price discounts using
  semiparametric regression}.
\newblock \bibinfo{journal}{\emph{Journal of Product \& Brand Management}}
  (\bibinfo{year}{2006}).
\newblock


\bibitem[\protect\citeauthoryear{Morgan and Winship}{Morgan and
  Winship}{2015}]%
        {morgan2015counterfactuals}
\bibfield{author}{\bibinfo{person}{Stephen~L Morgan} {and}
  \bibinfo{person}{Christopher Winship}.} \bibinfo{year}{2015}\natexlab{}.
\newblock \bibinfo{booktitle}{\emph{Counterfactuals and causal inference}}.
\newblock \bibinfo{publisher}{Cambridge University Press}.
\newblock


\bibitem[\protect\citeauthoryear{Pearl}{Pearl}{2009}]%
        {pearl2009causality}
\bibfield{author}{\bibinfo{person}{Judea Pearl}.}
  \bibinfo{year}{2009}\natexlab{}.
\newblock \bibinfo{booktitle}{\emph{Causality}}.
\newblock \bibinfo{publisher}{Cambridge university press}.
\newblock


\bibitem[\protect\citeauthoryear{Pearl et~al\mbox{.}}{Pearl
  et~al\mbox{.}}{2009}]%
        {pearl2009causal}
\bibfield{author}{\bibinfo{person}{Judea Pearl} {et~al\mbox{.}}}
  \bibinfo{year}{2009}\natexlab{}.
\newblock \showarticletitle{Causal inference in statistics: An overview}.
\newblock \bibinfo{journal}{\emph{Statistics surveys}}  \bibinfo{volume}{3}
  (\bibinfo{year}{2009}), \bibinfo{pages}{96--146}.
\newblock


\bibitem[\protect\citeauthoryear{Perakis and Sood}{Perakis and Sood}{2006}]%
        {perakis2006competitive}
\bibfield{author}{\bibinfo{person}{Georgia Perakis} {and}
  \bibinfo{person}{Anshul Sood}.} \bibinfo{year}{2006}\natexlab{}.
\newblock \showarticletitle{Competitive multi-period pricing for perishable
  products: A robust optimization approach}.
\newblock \bibinfo{journal}{\emph{Mathematical Programming}}
  \bibinfo{volume}{107}, \bibinfo{number}{1-2} (\bibinfo{year}{2006}),
  \bibinfo{pages}{295--335}.
\newblock


\bibitem[\protect\citeauthoryear{Phillips}{Phillips}{2005}]%
        {phillips2005pricing}
\bibfield{author}{\bibinfo{person}{Robert~Lewis Phillips}.}
  \bibinfo{year}{2005}\natexlab{}.
\newblock \bibinfo{booktitle}{\emph{Pricing and revenue optimization}}.
\newblock \bibinfo{publisher}{Stanford University Press}.
\newblock


\bibitem[\protect\citeauthoryear{Prosperi, Guo, Sperrin, Koopman, Min, He,
  Rich, Wang, Buchan, and Bian}{Prosperi et~al\mbox{.}}{2020}]%
        {prosperi2020causal}
\bibfield{author}{\bibinfo{person}{Mattia Prosperi}, \bibinfo{person}{Yi Guo},
  \bibinfo{person}{Matt Sperrin}, \bibinfo{person}{James~S Koopman},
  \bibinfo{person}{Jae~S Min}, \bibinfo{person}{Xing He},
  \bibinfo{person}{Shannan Rich}, \bibinfo{person}{Mo Wang},
  \bibinfo{person}{Iain~E Buchan}, {and} \bibinfo{person}{Jiang Bian}.}
  \bibinfo{year}{2020}\natexlab{}.
\newblock \showarticletitle{Causal inference and counterfactual prediction in
  machine learning for actionable healthcare}.
\newblock \bibinfo{journal}{\emph{Nature Machine Intelligence}}
  \bibinfo{volume}{2}, \bibinfo{number}{7} (\bibinfo{year}{2020}),
  \bibinfo{pages}{369--375}.
\newblock


\bibitem[\protect\citeauthoryear{Robins, Hernan, and Brumback}{Robins
  et~al\mbox{.}}{2000}]%
        {robins2000marginal}
\bibfield{author}{\bibinfo{person}{James~M Robins},
  \bibinfo{person}{Miguel~Angel Hernan}, {and} \bibinfo{person}{Babette
  Brumback}.} \bibinfo{year}{2000}\natexlab{}.
\newblock \bibinfo{title}{Marginal structural models and causal inference in
  epidemiology}.
\newblock
\newblock


\bibitem[\protect\citeauthoryear{Shalit, Johansson, and Sontag}{Shalit
  et~al\mbox{.}}{2017}]%
        {shalit2017estimating}
\bibfield{author}{\bibinfo{person}{Uri Shalit}, \bibinfo{person}{Fredrik~D
  Johansson}, {and} \bibinfo{person}{David Sontag}.}
  \bibinfo{year}{2017}\natexlab{}.
\newblock \showarticletitle{Estimating individual treatment effect:
  generalization bounds and algorithms}. In
  \bibinfo{booktitle}{\emph{International Conference on Machine Learning}}.
  PMLR, \bibinfo{pages}{3076--3085}.
\newblock


\bibitem[\protect\citeauthoryear{van Ryzin}{van Ryzin}{2005}]%
        {van2005models}
\bibfield{author}{\bibinfo{person}{Garrett~J van Ryzin}.}
  \bibinfo{year}{2005}\natexlab{}.
\newblock \showarticletitle{Models of demand}.
\newblock In \bibinfo{booktitle}{\emph{The Oxford Handbook of Pricing
  Management}}.
\newblock


\bibitem[\protect\citeauthoryear{Wu, Li, and Da~Xu}{Wu et~al\mbox{.}}{2014}]%
        {wu2014randomized}
\bibfield{author}{\bibinfo{person}{Jianghua Wu}, \bibinfo{person}{Ling Li},
  {and} \bibinfo{person}{Li Da~Xu}.} \bibinfo{year}{2014}\natexlab{}.
\newblock \showarticletitle{A randomized pricing decision support system in
  electronic commerce}.
\newblock \bibinfo{journal}{\emph{Decision Support Systems}}
  \bibinfo{volume}{58} (\bibinfo{year}{2014}), \bibinfo{pages}{43--52}.
\newblock


\bibitem[\protect\citeauthoryear{Yabe, Ito, and Fujimaki}{Yabe
  et~al\mbox{.}}{2017}]%
        {yabe2017robust}
\bibfield{author}{\bibinfo{person}{Akihiro Yabe}, \bibinfo{person}{Shinji Ito},
  {and} \bibinfo{person}{Ryohei Fujimaki}.} \bibinfo{year}{2017}\natexlab{}.
\newblock \showarticletitle{Robust Quadratic Programming for Price
  Optimization.}. In \bibinfo{booktitle}{\emph{IJCAI}}.
  \bibinfo{pages}{4648--4654}.
\newblock


\bibitem[\protect\citeauthoryear{Zhang et~al\mbox{.}}{Zhang
  et~al\mbox{.}}{2006}]%
        {zhang2006multi}
\bibfield{author}{\bibinfo{person}{Lei Zhang} {et~al\mbox{.}}}
  \bibinfo{year}{2006}\natexlab{}.
\newblock \emph{\bibinfo{title}{Multi-period pricing for perishable products:
  Uncertainty and competition}}.
\newblock \bibinfo{thesistype}{Ph.D. Dissertation}.
  \bibinfo{school}{Massachusetts Institute of Technology}.
\newblock


\bibitem[\protect\citeauthoryear{Zhao, Hua, Yan, Zhang, Xu, and Yang}{Zhao
  et~al\mbox{.}}{2019}]%
        {zhao2019unified}
\bibfield{author}{\bibinfo{person}{Kui Zhao}, \bibinfo{person}{Junhao Hua},
  \bibinfo{person}{Ling Yan}, \bibinfo{person}{Qi Zhang}, \bibinfo{person}{Huan
  Xu}, {and} \bibinfo{person}{Cheng Yang}.} \bibinfo{year}{2019}\natexlab{}.
\newblock \showarticletitle{A Unified Framework for Marketing Budget
  Allocation}. In \bibinfo{booktitle}{\emph{Proceedings of the 25th ACM SIGKDD
  International Conference on Knowledge Discovery \& Data Mining}}.
  \bibinfo{pages}{1820--1830}.
\newblock


\end{thebibliography}

\newpage

\appendix

\section{SUPPLEMENT}

\subsection{Implementation} 
In order to improve the reproducibility, we present more details about the implementation. 
\subsubsection{Software versions}
The details of programming languages, software packages and frameworks used in the experiments and deployed system are as follow:
\begin{itemize}
	\item Language: Python 3.6.10, SQL. 
	\item Packages: XGoost 1.3.1, NumPy 1.16.6, Pandas 1.0.3, Tensorflow 1.14.0, DeepIV\footnote{https://github.com/jhartford/DeepIV}.
	\item Frameworks: MaxCompute.
\end{itemize}

\subsubsection{Feature extraction} We first extract the features of both products and stores. The raw features includes brand, sku, department, categories,  weekend, holiday information, sales channel,  promotional event, page views,  users views, historical discounts and historical sales, etc. Secondly, we use the data aggregation process to create new features.  Specifically, we aggregate the sales with different time period, such as by week and by holiday, and aggregate the sales with different clusters, such brand, category, store, sku, channel of sales.
Thirdly, we impute the missing data by using the averaged value or by nearest neighbor matching.  Other methods can also be applied, such as EM algorithm, interpolation and matrix completion, etc.  Finally, one-hot representation is carried out for the sparse features, such as one-level, second-level, and third-level categories.  

\subsubsection{Other tips}
\begin{itemize}
	\item  The normalized factor $Y_{i}^{nor}$ is the average sales of product $i$ by definition. However, this quantity may be not very stable.  In practice, the average sales of level-2 or level-3 category that the product belongs to can be used for normalization. 
	\item The division operation is sensitive to the small value. It is better to add a small quantity before do division operation, e.g., $\frac{a}{b} \to \frac{a}{b+1}$.
\end{itemize}

\end{document}